%% file: main.tex
\documentclass{article}

\input{packages}
\input{notation}

\input{editor}

\usepackage[final,nonatbib]{neurips_2021}
\usepackage[numbers,sort]{natbib}

\usepackage[utf8]{inputenc} % allow utf-8 input
\usepackage[T1]{fontenc}    % use 8-bit T1 fonts
\usepackage{hyperref}       % hyperlinks
\usepackage{url}            % simple URL typesetting
\usepackage{booktabs}       % professional-quality tables
\usepackage{amsfonts}       % blackboard math symbols
\usepackage{nicefrac}       % compact symbols for 1/2, etc.
\usepackage{microtype}      % microtypography
\usepackage{xcolor}         % colors

\newcommand{\dproc}{d_\text{Proc}}

\title{Grounding Representation Similarity with Statistical Testing}

\author{%
  Frances Ding, \hspace{.25cm} Jean-Stanislas Denain, \hspace{.25cm} Jacob Steinhardt
  \\
  University of California Berkeley\\
  \texttt{\{frances, js\_denain, jsteinhardt\}@berkeley.edu} \\
}

\begin{document}

\maketitle

\begin{abstract}
	\input{abstract}
\end{abstract}

\section{Introduction}

\input{intro}

\section{Problem Setup: Metrics and Models} \label{section:setup}
\input{setup}

\section{Warm-up: Intuitive Tests for Sensitivity and Specificity} \label{section:disagreement}
\input{metric_disagreement}

\section{Rigorously Evaluating Dissimilarity Metrics} \label{section:results}
\input{results}

%\section{CSA?}
\section{Discussion} \label{section:discussion}
\input{discussion}

%\subsection{Tables}
%
%All tables must be centered, neat, clean and legible.  The table number and
%title always appear before the table.  See Table~\ref{sample-table}.
%
%Place one line space before the table title, one line space after the
%table title, and one line space after the table. The table title must
%be lower case (except for first word and proper nouns); tables are
%numbered consecutively.
%
%Note that publication-quality tables \emph{do not contain vertical rules.} We
%strongly suggest the use of the \verb+booktabs+ package, which allows for
%typesetting high-quality, professional tables:
%\begin{center}
%  \url{https://www.ctan.org/pkg/booktabs}
%\end{center}
%This package was used to typeset Table~\ref{sample-table}.
%
%\begin{table}
%  \caption{Sample table title}
%  \label{sample-table}
%  \centering
%  \begin{tabular}{lll}
%    \toprule
%    \multicolumn{2}{c}{Part}                   \\
%    \cmidrule(r){1-2}
%    Name     & Description     & Size ($\mu$m) \\
%    \midrule
%    Dendrite & Input terminal  & $\sim$100     \\
%    Axon     & Output terminal & $\sim$10      \\
%    Soma     & Cell body       & up to $10^6$  \\
%    \bottomrule
%  \end{tabular}
%\end{table}
\clearpage

\begin{ack}
Thanks to Ruiqi Zhong for helpful comments and assistance in finetuning models, and thanks to Daniel Rothchild and our anonymous reviewers for helpful discussion.
FD is supported by the National Science Foundation Graduate Research Fellowship Program under Grant No. DGE 1752814 and the Open Philanthropy Project AI Fellows Program. JSD is supported by the NSF Division of Mathematical Sciences Grant No. 2031985.

%Use unnumbered first level headings for the acknowledgments. All acknowledgments
%go at the end of the paper before the list of references. Moreover, you are required to declare
%funding (financial activities supporting the submitted work) and competing interests (related financial activities outside the submitted work).
%More information about this disclosure can be found at: \url{https://neurips.cc/Conferences/2021/PaperInformation/FundingDisclosure}.

%Do {\bf not} include this section in the anonymized submission, only in the final paper. You can use the \texttt{ack} environment provided in the style file to autmoatically hide this section in the anonymized submission.
\end{ack}

%\section*{References}
%
%References follow the acknowledgments. Use unnumbered first-level heading for
%the references. Any choice of citation style is acceptable as long as you are
%consistent. It is permissible to reduce the font size to \verb+small+ (9 point)
%when listing the references.
%Note that the Reference section does not count towards the page limit.
\medskip

\bibliographystyle{abbrvnat}

\bibliography{mybib}

\clearpage

\newpage

%%%%%%%%%%%%%%%%%%%%%%%%%%%%%%%%%%%%%%%%%%%%%%%%%%%%%%%%%%%%

%\input{checklist}

%%%%%%%%%%%%%%%%%%%%%%%%%%%%%%%%%%%%%%%%%%%%%%%%%%%%%%%%%%%%
\clearpage
\appendix

\input{appendix}

\end{document}

%% file: packages.tex
\usepackage{booktabs} % For formal tables
\usepackage{xfrac}
\usepackage{algorithm}
\usepackage[noend]{algorithmic}
\usepackage{amsmath,amsthm,amsfonts}
\usepackage{amsmath}
\usepackage{color}
\usepackage{mathrsfs}
\usepackage{verbatim}
\usepackage{enumitem}
\usepackage{bm}
\usepackage{hyperref}
\usepackage{xr}
\usepackage{tikz}
\usepackage{caption}
\usepackage{subcaption}
\usepackage{float}
\usepackage{listings}
\usepackage{array}
\usepackage{multirow}
\usepackage{makecell}
\usepackage{mathtools}
\usepackage{cases}

\definecolor{DarkBlue}{rgb}{0.1,0.1,0.5}
\hypersetup{
	colorlinks=true,       % false: boxed links; true: colored links
	linkcolor=DarkBlue,          % color of internal links
	citecolor=DarkBlue,        % color of links to bibliography
	filecolor=DarkBlue,      % color of file links
	urlcolor=DarkBlue,          % color of external links
	pdftitle={},
	pdfauthor={},
}

%% file: notation.tex
\newcommand{\R}{\mathbb{R}}

\newtheorem*{proposition*}{Proposition}

%% file: editor.tex
\usepackage[normalem]{ulem}
\usepackage{xcolor}

\newcommand{\scrcmd}{\textsuperscript}
\newcommand{\scr}[1]{\scrcmd{#1}}

%% file: abstract.tex
To understand neural network behavior, recent works quantitatively compare different networks' learned representations using canonical correlation analysis (CCA), centered kernel alignment (CKA), and other dissimilarity measures. Unfortunately, these widely used measures often disagree on fundamental observations, such as whether deep networks differing only in random initialization learn similar representations. These disagreements raise the question: which, if any, of these dissimilarity measures should we believe? We provide a framework to ground this question through a concrete test: measures should have \emph{sensitivity} to changes that affect functional behavior, and \emph{specificity} against changes that do not. We quantify this through a variety of functional behaviors including probing accuracy and robustness to distribution shift, and examine changes such as varying random initialization and deleting principal components. We find that current metrics exhibit different weaknesses, note that a classical baseline performs surprisingly well, and highlight settings where all metrics appear to fail, thus providing a challenge set for further improvement.

%% file: intro.tex
Understanding neural networks is not only scientifically interesting, but critical for applying deep networks in high-stakes situations. Recent work has highlighted the value of analyzing not just the final outputs of a network, but also its intermediate representations \citep{li2015convergent, raghu2019rapid}. This has motivated the development of representation similarity measures, which can provide insight into how different training schemes, architectures, and datasets affect networks' learned representations.

A number of similarity measures have been proposed, including centered kernel alignment (CKA) \citep{kornblith2019similarity}, ones based on canonical correlation analysis (CCA) \citep{raghu2017svcca, morcos2018insights}, single neuron alignment \citep{li2015convergent}, vector space alignment \citep{arora2017asimple, smith2017offline, conneau2018word}, and others \citep{laakso2000content, NEURIPS2018_5fc34ed3, liang2019knowledge, lenc2015understanding, alain2018understanding, feng2020transferred}. Unfortunately, these different measures tell different stories. For instance, CKA and projection weighted CCA disagree on which layers 
of different networks are most similar \citep{kornblith2019similarity}. 
This lack of consensus is worrying, as measures are often designed according to different and incompatible intuitive desiderata, such as whether finding a one-to-one assignment, or finding few-to-one mappings, between neurons is more appropriate \citep{li2015convergent}. 
As a community, we need well-chosen formal criteria for evaluating metrics to avoid over-reliance on intuition and the pitfalls of too many researcher degrees of freedom \citep{leavitt2020towards}.

In this paper we view representation dissimilarity measures as implicitly answering a classification question--whether two representations are essentially similar or importantly different. 
Thus, in analogy to statistical testing, we can evaluate them based on their \emph{sensitivity} to important change and \emph{specificity} (non-responsiveness) 
against unimportant changes or noise. 

As a warm-up, we first initially consider two intuitive criteria: 
first, that metrics should have specificity against random initialization; and second, that they should be sensitive to deleting 
important principal components (those that affect probing accuracy). Unfortunately, popular 
metrics fail at least one of these two tests. CCA is not specific -- random initialization noise overwhelms differences between even far-apart 
layers in a network (Section \ref{section:disagreement-layer}). CKA on the other hand is not sensitive, failing to detect changes in all but the top $10$ principal components 
of a representation (Section \ref{section:disagreement-pca}). 

We next construct quantitative benchmarks to evaluate a dissimilarity measure's quality. To move beyond our intuitive 
criteria, we need a ground truth. For this we turn to the functional behavior of the representations we are comparing, 
measured through probing accuracy (an indicator of syntactic information) \citep{belinkov2017neural, peters2018dissecting, tenney2018what}
and out-of-distribution performance of the model they belong to \citep{naik2018stress, mccoy2020berts, d2020underspecification}. 
We then score dissimilarity measures based on their rank correlation with these measured functional differences. Overall our 
benchmarks contain 30,480 examples and vary representations across several axes including random seed, layer depth, and 
low-rank approximation (Section~\ref{section:results})\footnote{Code to replicate our results can be found at \url{https://github.com/js-d/sim_metric}.}.

Our benchmarks confirm our two intuitive observations: on subtasks that consider layer depth and principal component deletion, we measure the rank correlation with 
probing accuracy and find CCA and CKA lacking as the previous warm-up experiments suggested.  
Meanwhile, the Orthogonal Procrustes distance, a classical but often overlooked\footnote{For instance, \citet{raghu2017svcca} and \citet{morcos2018insights} do not mention it, and \citet{kornblith2019similarity} relegates it to the appendix; 
although \citet{smith2017offline} does use it to analyze word embeddings and prefers it to CCA.}
dissimilarity measure, balances gracefully between CKA and CCA and consistently performs well. 
This underscores the need for systematic evaluation, otherwise we may fall to recency bias that undervalues classical baselines.

Other subtasks measure correlation with OOD accuracy, motivated by the observation that random initialization 
sometimes has large effects on OOD performance \citep{mccoy2020berts}. We find that dissimilarity measures can 
sometimes predict OOD performance using only the in-distribution representations, but we also identify a challenge 
set on which none of the measures do statistically better than chance.
We hope this challenge set will help measure and spur progress in the future.

%% file: setup.tex
Our goal is to quantify the similarity between two different groups of neurons (usually layers). 
We do this by comparing how their activations behave on the same dataset. 
Thus for a layer with $p_1$ neurons, we define $A \in \mathbb{R}^{p_1 \times n}$, the matrix of activations of the $p_1$ neurons on $n$ data points, to be that layer's raw representation of the data. 
Similarly, let $B \in \R^{p_2 \times n}$ be a matrix of the activations of $p_2$ neurons on the same $n$ data points. 
We center and normalize these representations before computing dissimilarity, per standard practice. 
Specifically, for a raw representation $A$ we first subtract the mean value from each column, 
then divide by the Frobenius norm, 
to produce the normalized representation $A^*$, used in all our dissimilarity computations. 
In this work we study dissimilarity measures $d(A^*, B^*)$ that allow for quantitative comparisons of representations both within and across different networks. 
We colloquially refer to values of $d(A^*,B^*)$ as distances, although they do not necessarily satisfy the triangle inequality required of a proper metric.  

We study five dissimilarity measures: centered kernel alignment (CKA), three measures derived from canonical correlation analysis (CCA), and a measure derived from the orthogonal Procrustes problem.
%Previous work has argued that similarity measures should have certain properties, such as invariance to left orthogonal transformations to accommodate the symmetries of neural networks \citet{kornblith2019similarity} and all measures except for projection-weighted CCA satisfy this requirement.

\textbf{Centered kernel alignment (CKA)} uses an inner product to quantify similarity between two representations. It is based on the idea that one can first choose a kernel, compute the $n \times n$ kernel matrix for each representation, and then measure similarity as the alignment between these two kernel matrices.  
The measure of similarity thus depends on one's choice of kernel; in this work we consider \textbf{Linear CKA}:
\begin{equation}
    d_\text{Linear CKA}(A, B) = 1 - \frac{\|AB^\top\|_F^2}{\|AA^\top\|_F \|BB^\top\|_F}
\end{equation}
as proposed in \citet{kornblith2019similarity}. Other choices of kernel are also valid; we focus on Linear CKA here since \citet{kornblith2019similarity} report similar results from using either a linear or RBF kernel.  

\textbf{Canonical correlation analysis (CCA)} finds orthogonal bases ($w_A^i, w_B^i$) for two matrices such that after projection onto $w_A^i, w_B^i$, the projected matrices have maximally correlated rows. For $1 \leq i \leq p_1$, the $i^\text{th}$ canonical correlation coefficient $\rho_i$ is computed as follows:
\begin{align}
    \rho_i &= \max_{w_A^i, w_B^i} \frac{\langle {w_A^i}^\top A, {w_B^i}^\top B\rangle}{\|{w_A^i}^\top A\| \cdot \|{w_B^i}^\top B\| }\\
    s.t.~\hspace{.2cm} \langle {w_A^i}^\top A , {w_A^j}^\top A \rangle &= 0, ~\forall j < i, \hspace{.5cm}
    \langle {w_B^i}^\top B,  {w_B^j}^\top B \rangle = 0 , ~\forall j < i
\end{align}
To transform the vector of correlation coefficients into a scalar measure, two options considered previously \citep{kornblith2019similarity} are the \textbf{mean correlation coefficient, $\bar \rho_{\text{CCA}}$}, and the \textbf{mean squared correlation coefficient, $R^2_{\text{CCA}}$}, defined as follows:
\begin{align}
	d_{\bar \rho_{\text{CCA}}}(A, B) = 1 - \frac{1}{p_1} \sum_i \rho_i, \hspace{1cm} d_{R^2_{\text{CCA}}}(A, B) = 1 - \frac{1}{p_1} \sum_i \rho_i^2
\end{align}

To improve the robustness of CCA, \citet{morcos2018insights} propose \textbf{projection-weighted CCA (PWCCA)} as another scalar summary of CCA:
\begin{align}
    d_\text{PWCCA}(A, B) = 1 - \frac{\sum_i \alpha_i \rho_i}{\sum_i \alpha_i}, \hspace{.5cm} \alpha_i = \sum_j | \langle h_i, a_j \rangle| 
\end{align}
where $a_j$ is the $j^\text{th}$ row of $A$, and $h_i = {w_A^i}^\top A$ is the projection of $A$ onto the $i^\text{th}$ canonical direction.
We find that PWCCA performs far better than $\bar \rho_{\text{CCA}}$ and $R^2_{\text{CCA}}$, so we focus on PWCCA in the main text, but include results on the other two measures in the appendix.

The \textbf{orthogonal Procrustes} problem consists of finding the left-rotation of $A$ that is closest to $B$ in Frobenius norm, i.e. solving the optimization problem:
\begin{equation}
    \min\limits_R \|B - RA\|^2_{\text{F}}, \hspace{.2cm} \text{subject to } R^\top R = I.
\end{equation}

The minimum is the squared \textbf{orthogonal Procrustes distance} between \(A\) and \(B\), and is equal to 
\begin{equation}
    \dproc(A, B) = \|A\|_F^2 + \|B \|_F^2 - 2 \|A^\top B \|_*,
\end{equation}

where \(\|\cdot \|_*\) is the nuclear norm \citep{schonemann1966generalized}. Unlike the other metrics, the orthogonal Procrustes distance is not normalized between 0 and 1, although for normalized $A^*$, $B^*$ it lies in $[0, 2]$.

\subsection{Models we study}
In this work we study representations of both text and image inputs. 
For text, we investigate representations computed by Transformer architectures in the BERT model family \citep{devlin2018bert} on sentences from the Multigenre Natural Language Inference (MNLI) dataset \citep{N18-1101}. 
We study BERT models of two sizes: BERT base, with 12 hidden layers of 768 neurons, and BERT medium, with 8 hidden layers of 512 neurons. 
We use the same architectures as in the open source BERT release\footnote{available at \url{https://github.com/google-research/bert}}, but to generate diversity we study 3 variations of these models:
\begin{enumerate}
\setlength\itemsep{-.4em}
	\item $10$ BERT base models pretrained with different random seeds but not finetuned for particular tasks, released by \citet{zhong2021larger}\footnote{available at \url{https://github.com/ruiqi-zhong/acl2021-instance-level}}.
	\item $10$ BERT medium models initialized from pretrained models released by \citet{zhong2021larger}, that we further finetuned on MNLI with $10$ different finetuning seeds ($100$ models total).
	\item $100$ BERT base models that were initialized from the pretrained BERT model in \citep{devlin2018bert} and finetuned on MNLI with different seeds, released by \citet{mccoy2020berts}\footnote{available at \url{https://github.com/tommccoy1/hans/tree/master/berts_of_a_feather}}.
\end{enumerate}
 
For images, we investigate representations computed by ResNets \citep{he2016deep} on CIFAR-10 test set images \cite{krizhevsky2009learning}. We train $100$ ResNet-14 models\footnote{from \url{https://github.com/pytorch/vision/blob/master/torchvision/models/resnet.py}} from random initialization with different seeds on the CIFAR-10 training set and collect representations after each convolutional layer.
 
Further training details, as well as checks that our training protocols result in models with comparable performance to the original model releases, can be found in Appendix \ref{appendix:finetuning}.

%% file: metric_disagreement.tex
When designing dissimilarity measures, researchers usually consider invariants that these measures should not be sensitive to \citep{kornblith2019similarity}; 
for example, symmetries in neural networks imply that permuting the neurons in a fully connected layer does not change the representations learned. 
We take this one step further and frame dissimilarity measures as answering 
whether representations are essentially the same, or importantly different. 
We can then evaluate measures based on whether they respond to important changes 
(sensitivity) while ignoring changes that don't matter (specificity).

Assessing sensitivity and specificity requires a ground truth--which representations are truly different? To answer this, we begin with the following two intuitions\footnote{Note we will see later that these intuitions need refinement.}: 
1) neural network representations trained on the same data but from different random initializations are similar, and 
2) representations lose crucial information as principal components are deleted. 
These motivate the following intuitive tests of specificity and sensitivity: we expect a dissimilarity measure to: 
1) assign a small distance between architecturally identical neural networks that only differ in initialization seed, and 
2) assign a large distance between a representation $A$ and the representation $\hat A$ after deleting important principal components (enough to affect accuracy). 
We will see that PWCCA fails the first test (specificity), while CKA fails the second (sensitivity).

\input{heatmap_fig}

% Random seed vs. layer depth
\subsection{Specificity against changes to random seed} \label{section:disagreement-layer}

Neural networks with the same architecture trained from different random initializations show many similarities, such as highly correlated predictions on in-distribution data points \citep{mccoy2020berts}. 
Thus it seems natural to expect a good similarity measure to assign small distances between architecturally corresponding layers of networks that are identical except for initialization seed.

To check this property, we take two BERT base models pre-trained with different random seeds and, for every layer in the first model, compute its dissimilarity to every layer in both the first and second model. 
We do this for 5 separate pairs of models and average the results. 
To pass the intuitive specificity test, a dissimilarity measure should assign relatively small distances between a layer in the first network and its corresponding layer in the second network.

Figure \ref{fig:disagree-layer} displays the average pair-wise PWCCA, CKA, and Orthogonal Procrustes distances between layers of two networks differing only in random seed.
According to PWCCA, these networks' representations are quite dissimilar; 
for instance, the two layer $7$ representations are further apart than they are 
from any other layer in the same network.
PWCCA is thus not specific against random initialization, as it can outweigh even 
large changes in layer depth.

In contrast, CKA can separate layer $7$ in a different network from layers $4$ or $10$ in the same network, showing better specificity to random initialization. Orthogonal Procrustes exhibits smaller but non-trivial specificity, distinguishing 
layers once they are $4$-$5$ layers apart.

% PCA deletion
\subsection{Sensitivity to removing principal components} \label{section:disagreement-pca}

\begin{figure*}
    \centering
    
    \begin{subfigure}[b]{.47\linewidth}
    \includegraphics[width=\linewidth,trim={0 16cm 0 0},clip]{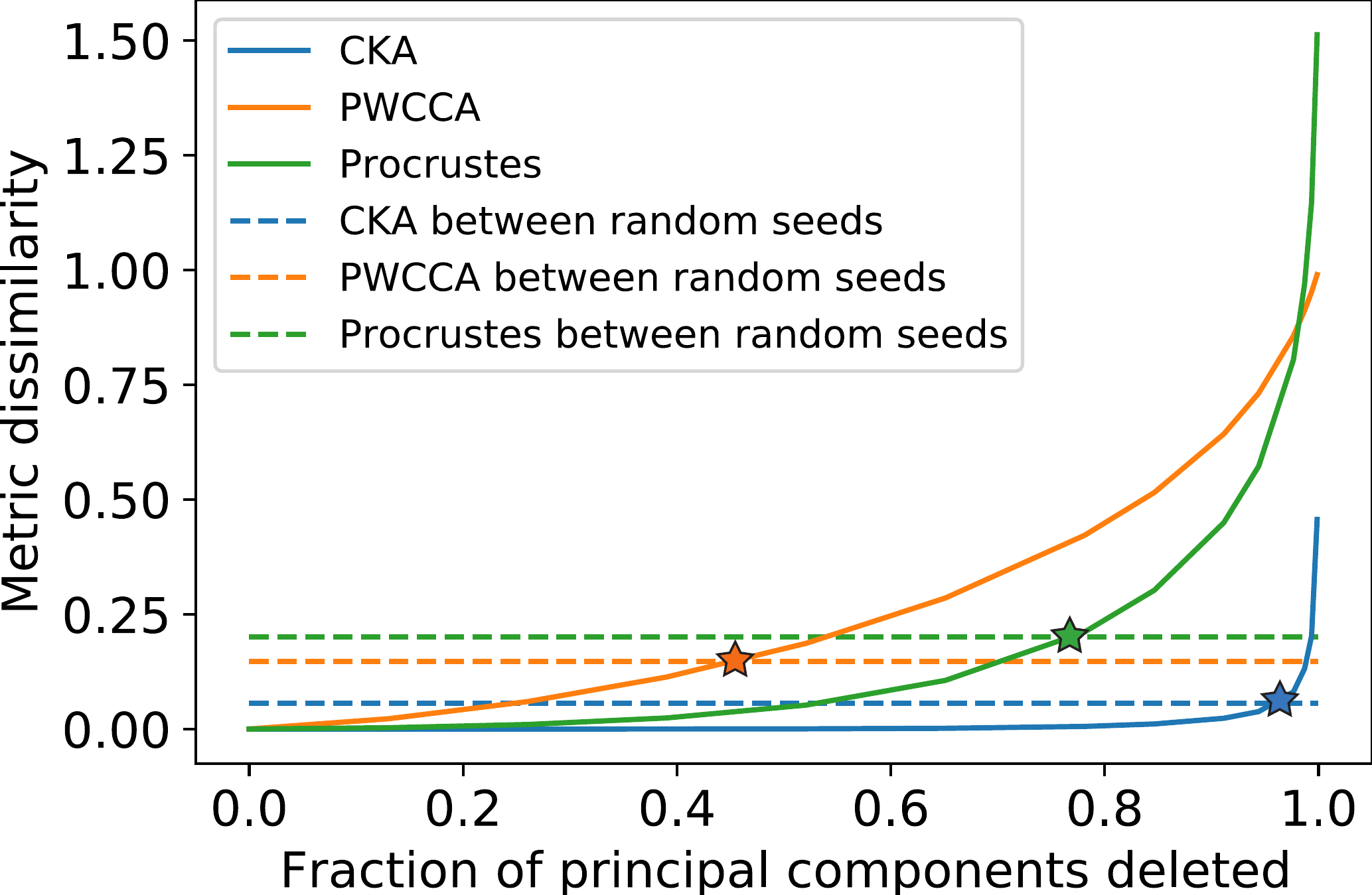}
    \caption{First layer of BERT}
    \end{subfigure}
    \hfill
    \begin{subfigure}[b]{0.47\linewidth}
    \includegraphics[width=\linewidth,trim={0 16cm 0 0},clip]{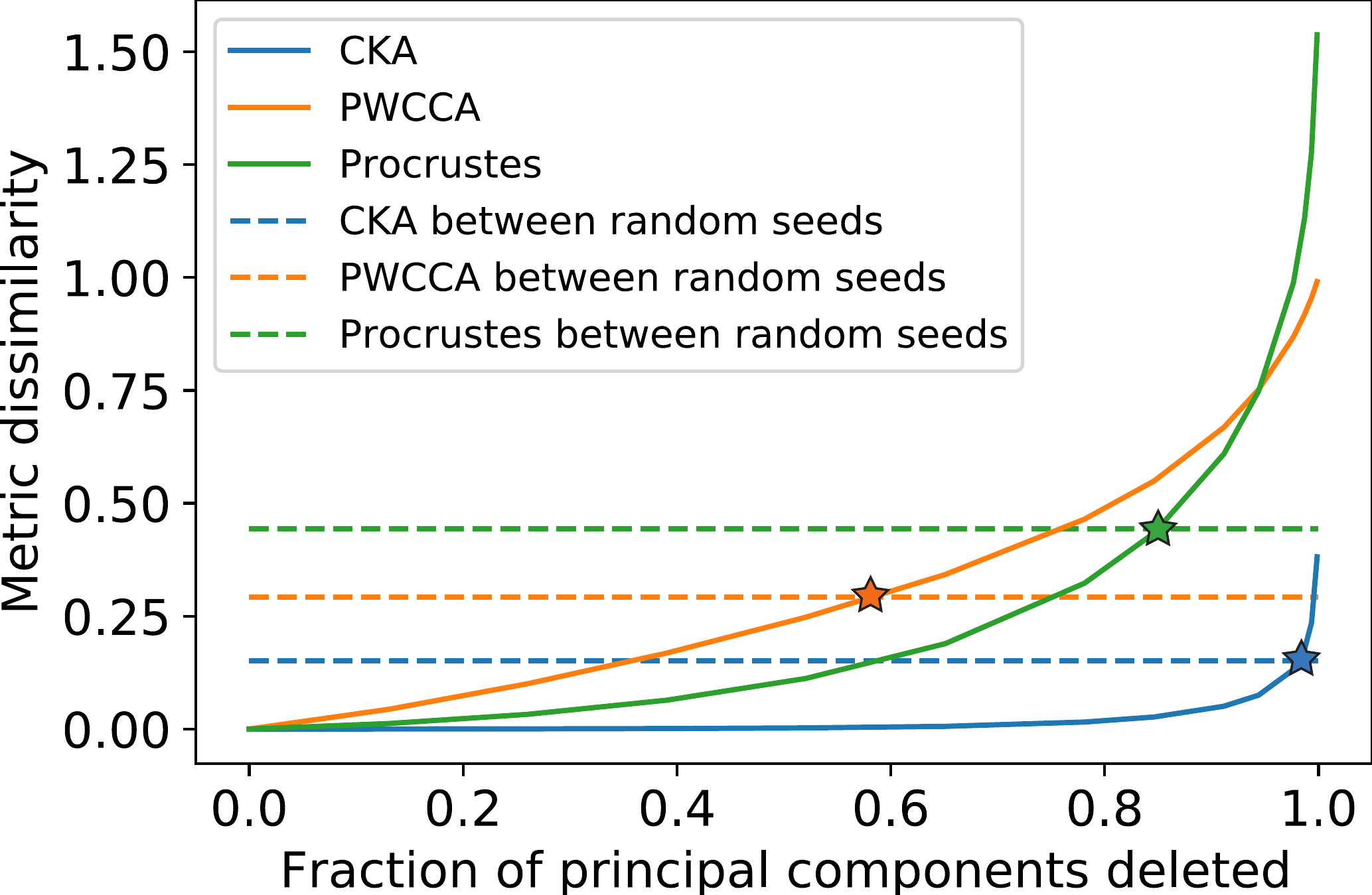}
    \caption{Last layer of BERT}
    \end{subfigure}
       \caption{\textbf{CKA fails to be sensitive to all but the largest principal components.} 
We compute dissimilarities between a layer's representation and low-rank approximations to that representation obtained by deleting principal components, starting from the smallest (solid lines). 
We also compute the average distance between networks trained with different random seeds as a 
baseline (dotted line), and mark the intersection point with a star. The starred points 
indicate that CKA requires almost all the components to be deleted before CKA distance exceeds 
the baseline.
}
    \label{fig:disagree-pca}
\end{figure*}

Dissimilarity measures should also be sensitive to deleting important principal components of a representation.% 
\footnote{For a representation $A$, we define $\hat A_{-k}$, the result of deleting the $k$ smallest principal components from $A$, as follows: we compute the singular value decomposition $U \Sigma V^T = A$, construct $U_{-k} \in \R^{p \times p-k}$ by dropping the lowest $k$ singular vectors of $U$, and finally take $\hat{A}_{-k} = U_{-k}^TA$.}
To quantify which components are important, we fix a layer of a pre-trained BERT base model and measure how probing accuracy degrades as principal components are deleted (starting from the smallest component), since probing accuracy is a common measure of the information captured in a representation \citep{belinkov2017neural}. 
We probe linear classification performance on the Stanford Sentiment Tree Bank task (SST-2) \citep{socher2013recursive}, 
following the experimental protocol in \citet{tamkin2020investigating}. Figure \ref{fig:var_pca_deletion} shows how probing accuracy degrades with component deletion.
Ideally, dissimilarity measures should be large by the time probing accuracy has decreased substantially.

To assess whether a dissimilarity measure is large, we need a baseline to compare to. For each measure, we define a dissimilarity score to be above the \emph{detectable} threshold if it is larger than the dissimilarity score between networks with different random initialization. 
Figure \ref{fig:disagree-pca} plots the dissimilarity induced by deleting principal components, 
as well as this baseline. 

For the last layer of BERT, CKA requires 97\% of a representation's principal components to be deleted for the dissimilarity to be detectable; after deleting these components, probing accuracy shown in Figure \ref{fig:var_pca_deletion} drops significantly from 80\% to 63\% (chance is $50\%$). 
CKA thus fails to detect large accuracy drops and so fails our intuitive sensitivity test. 

Other metrics perform better: Orthogonal Procrustes's detection threshold is $\sim$85\% of 
the principal components, corresponding to an accuracy drop 80\% to 70\%. 
PWCCA's threshold is $\sim$55\% of principal components, corresponding to an accuracy drop 
from 80\% to 75\%.

PWCCA's failure of specificity and CKA's failure of sensitivity on these intuitive tests are worrying. However, before declaring definitive failure, in the next section, we turn to making our assessments more rigorous.

%% file: heatmap_fig.tex
\begin{figure*}
	\centering
	\includegraphics[width=\linewidth,trim={0 21cm 0 0},clip]{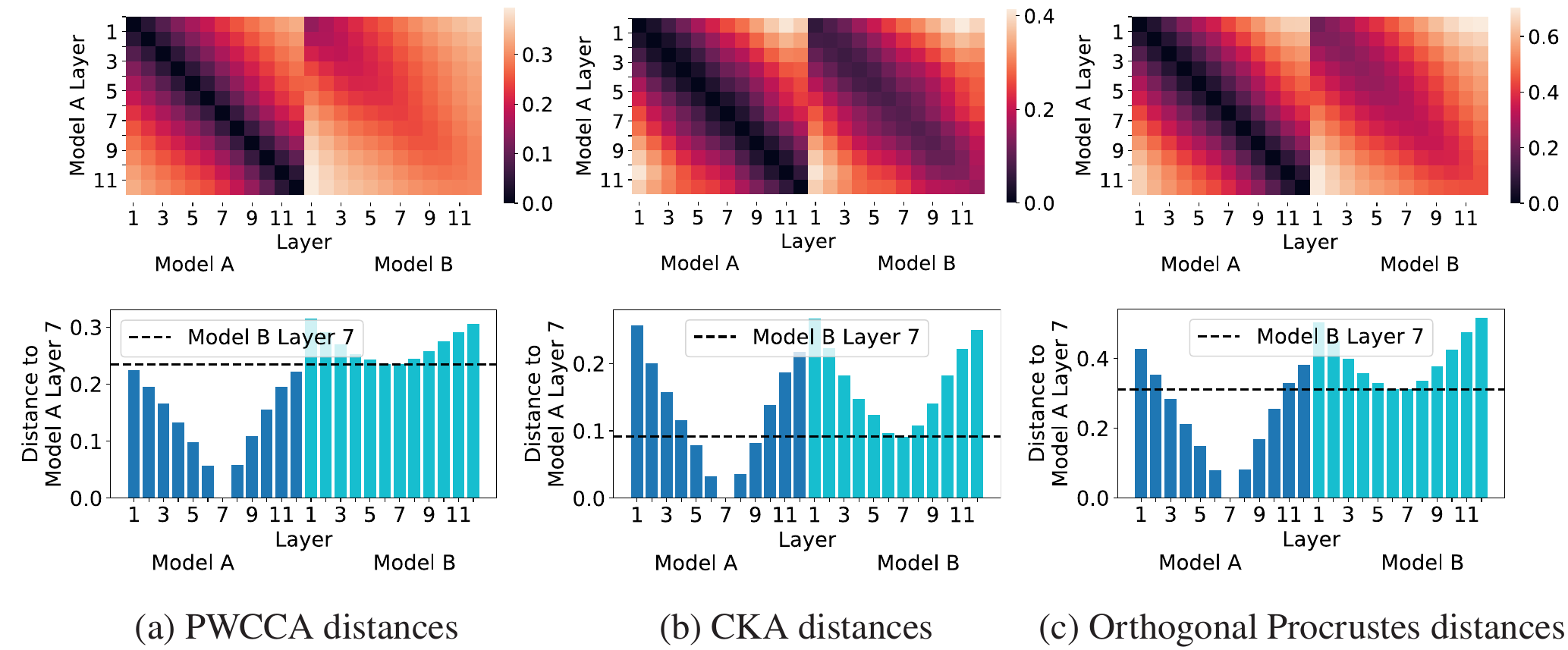}
	 \caption{\textbf{PWCCA fails the intuitive specificity test.} Top: PWCCA, CKA, and Orthogonal Procrustes pairwise distances between each layer of two differently initialized networks (Model A and B). Bottom: We zoom in to analyze the $7^\text{th}$ layer of Model A, plotting this layer's distance to every other layer in both networks; the dashed line indicates the distance 
to the corresponding $7^\text{th}$ layer in Model B. 
For PWCCA, none of the distances in model A exceed this line, 
indicating that random initialization affects this distance 
more than large changes in layer depth. }
	 \label{fig:disagree-layer}
\end{figure*}

%% file: results.tex
In the previous section, we saw that CKA and PWCCA each failed intuitive tests, based on sensitivity to principal components and 
specificity to random initialization. However, these were based primarily on intuitive, qualitative desiderata. Is there some way 
for us to make these tests more rigorous and quantitative?

\begin{figure*}
    \centering
    \begin{subfigure}[b]{0.48\linewidth}
		\includegraphics[width=1\linewidth]{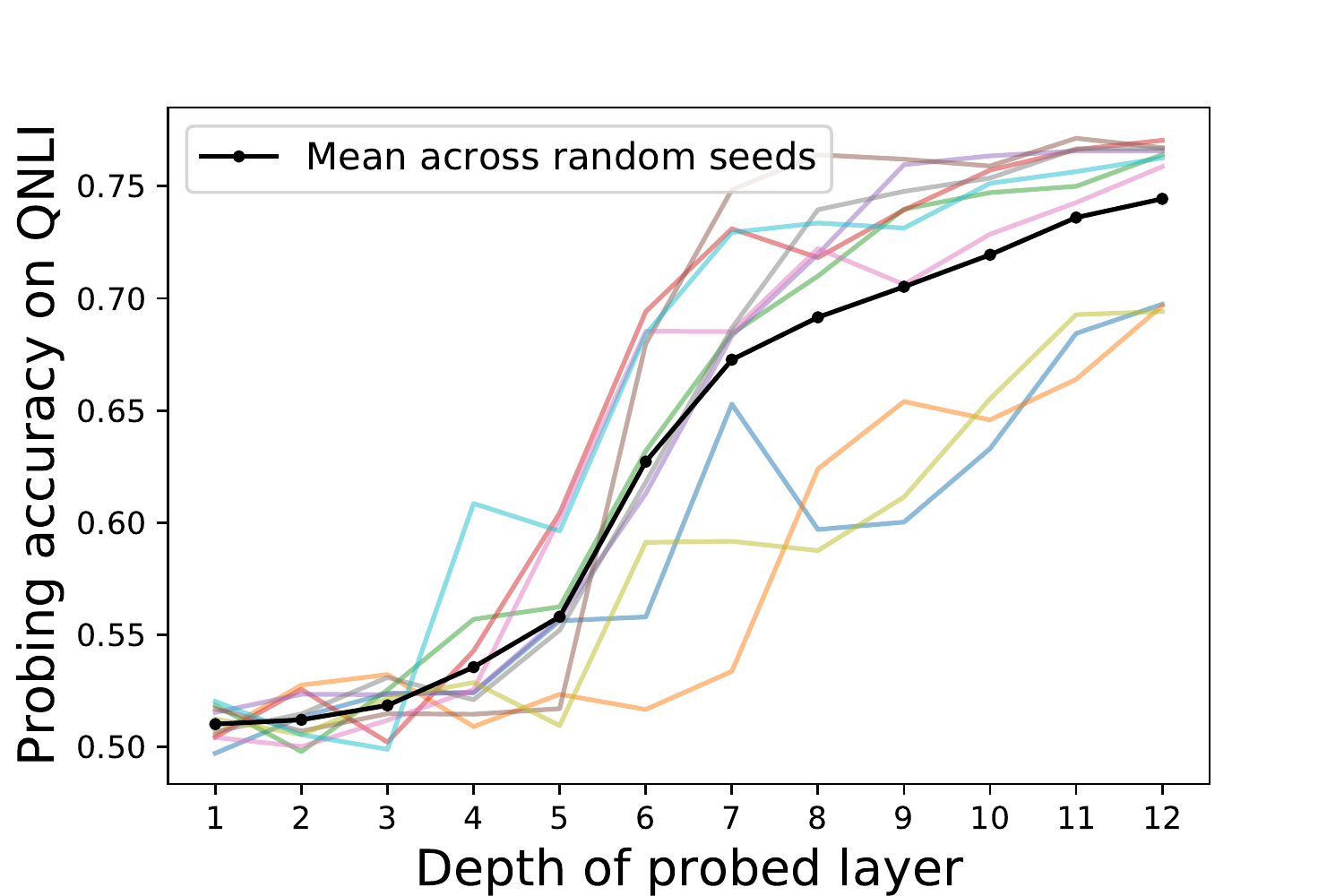}
		\caption{Probing variation across BERT base layers}
		\label{fig:var_layer_exp}
	\end{subfigure}
	\hfill
    \begin{subfigure}[b]{0.48\linewidth}
		\includegraphics[width=1\linewidth]{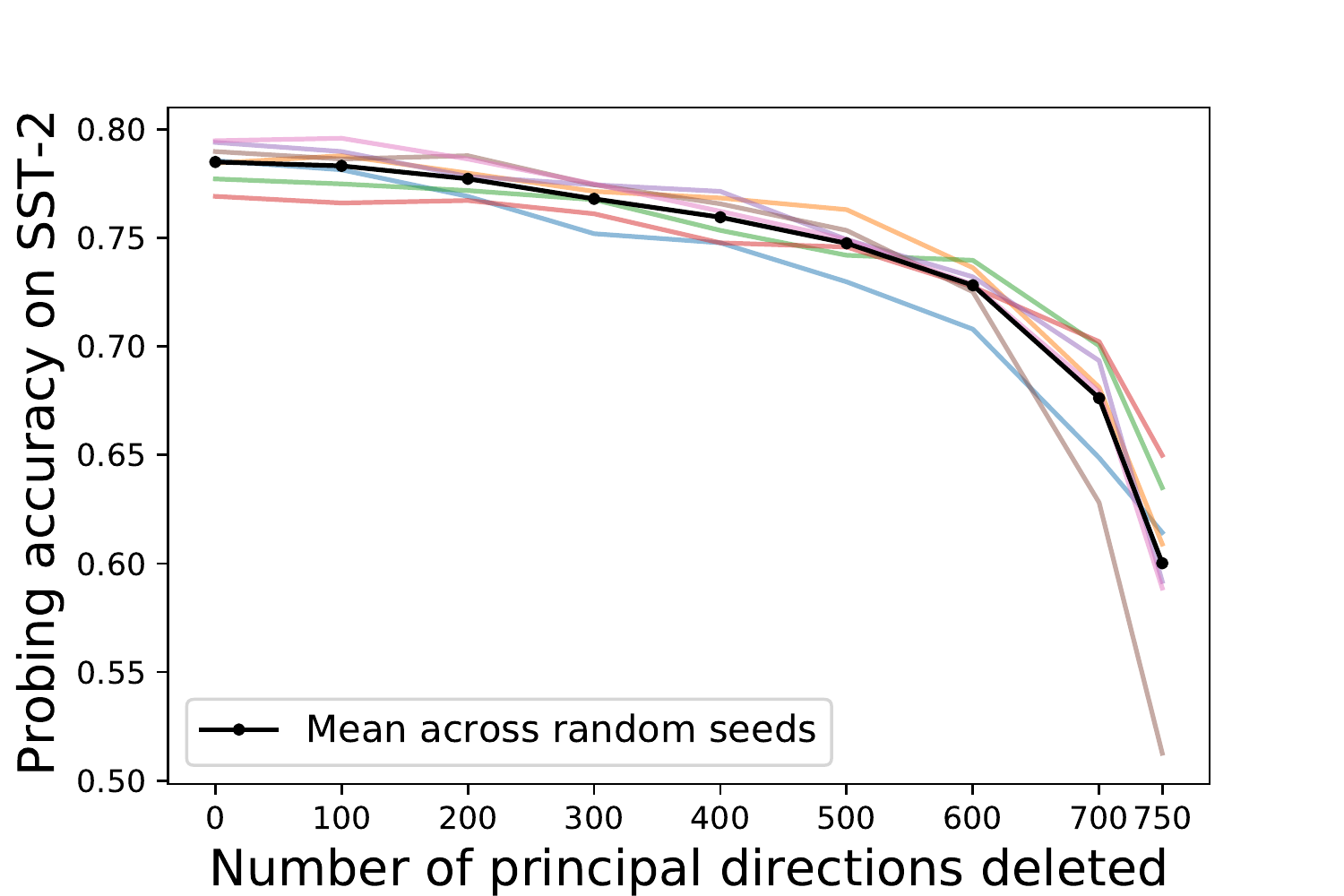}
		\caption{Variation across BERT base PC deletions}
		\label{fig:var_pca_deletion}
    \end{subfigure}
    \hfill
    \begin{subfigure}[b]{0.48\linewidth}
		\includegraphics[width=1\linewidth]{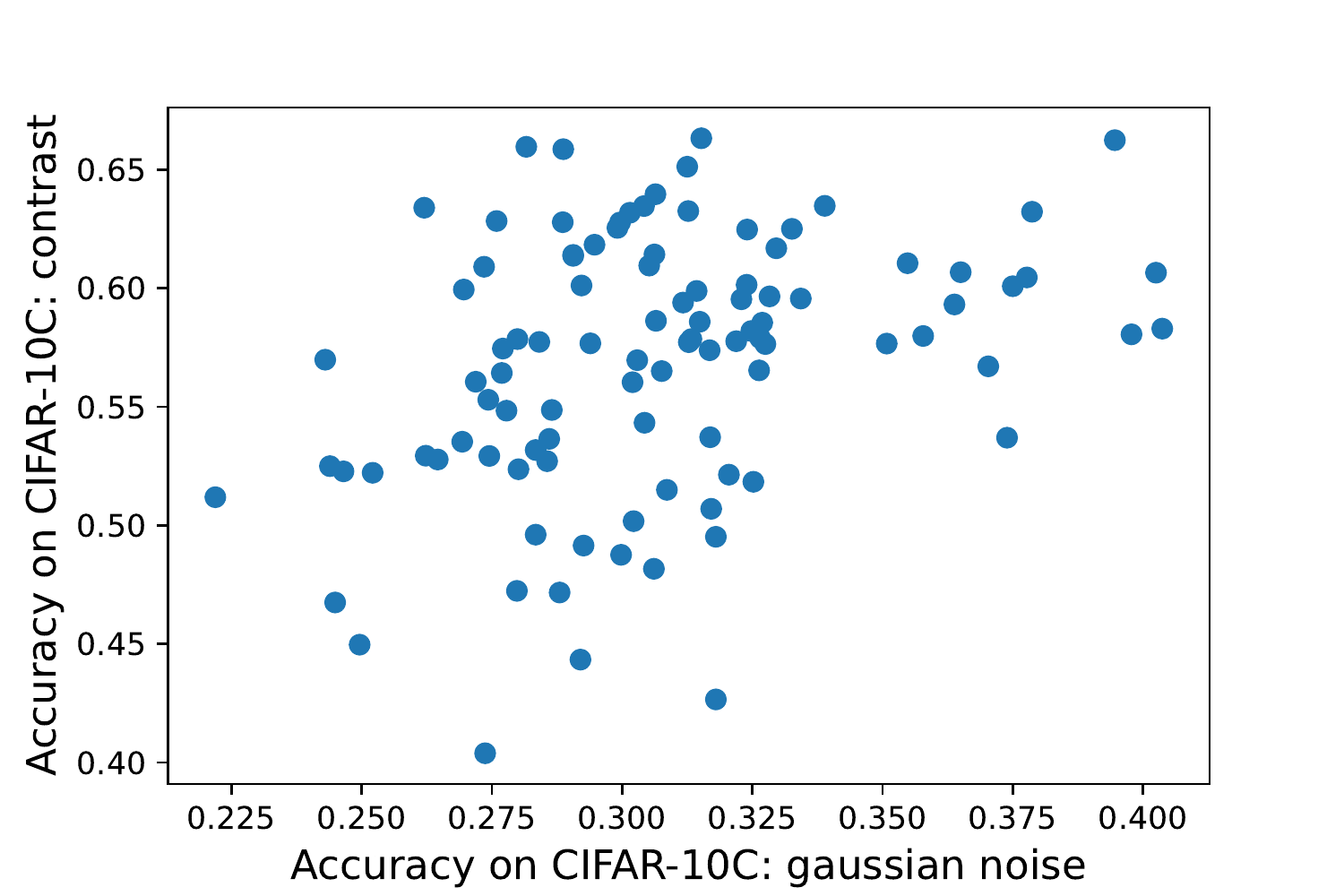}
		\caption{Variation across ResNet-14 training seeds}
		\label{fig:var_cifar10c}
    \end{subfigure}
    \hfill
    \begin{subfigure}[b]{0.48\linewidth}
		\includegraphics[width=1\linewidth,trim={0 15.7cm 0.2cm 0},clip]{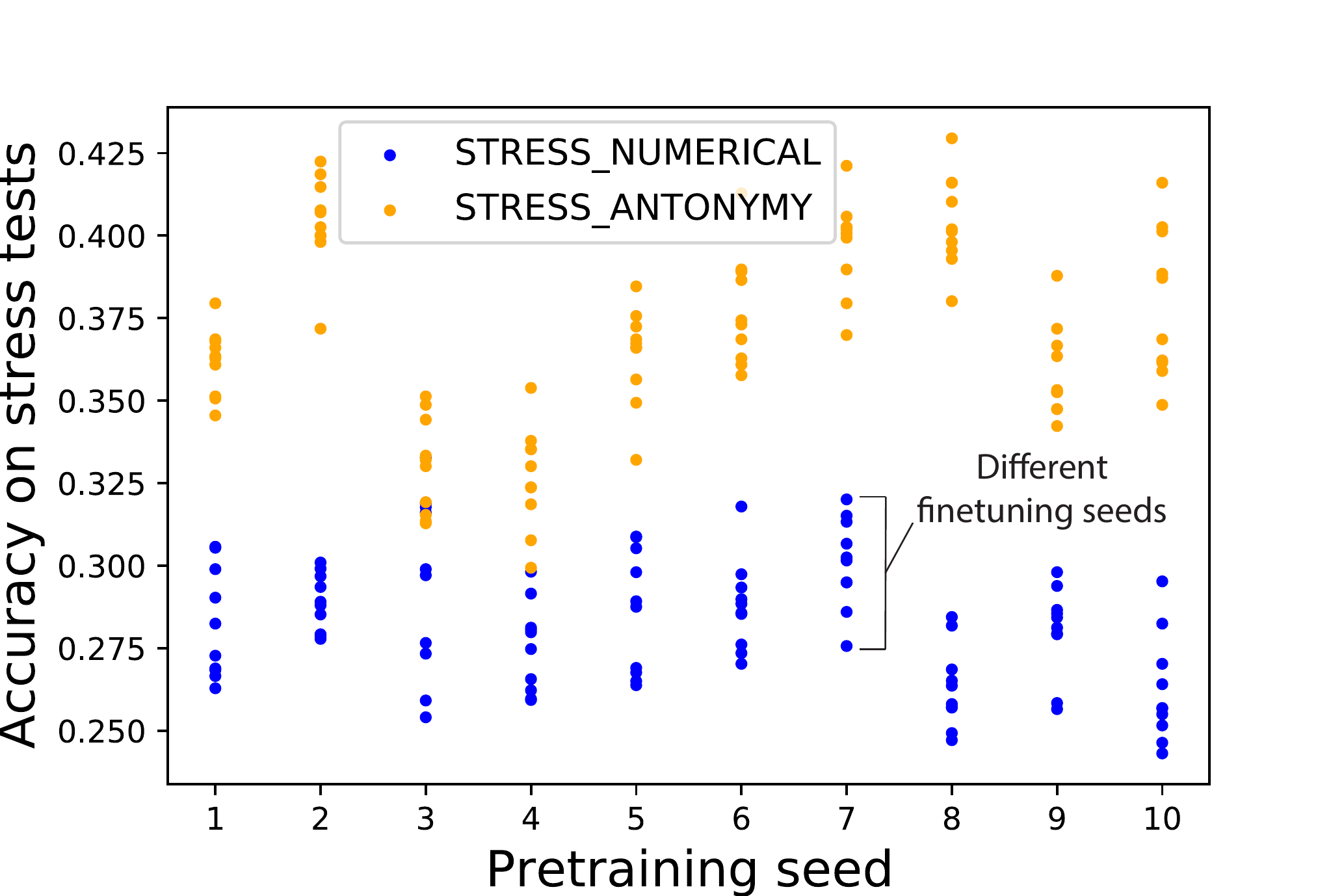}
		\caption{Variation across BERT medium training seeds}
		\label{fig:var_finetune_vs_finetune}
    \end{subfigure}
       \caption{\textbf{Our perturbations induce substantial variation on probing tasks and stress tests:} (\ref{fig:var_layer_exp}): Changing the depth of the examined BERT base layer strongly affects probing accuracy on QNLI. The trend for each randomly initialized model is displayed semi-transparently, and the solid black line is the mean trend. (\ref{fig:var_pca_deletion}): Truncating principal components from pretrained BERT base significantly degrades probing accuracy on SST-2 (BERT layer 12 shown here). (\ref{fig:var_cifar10c}): Training ResNet-14 on CIFAR-10 with different seeds leads to variation in accuracies on CIFAR-10C corruptions (here Gaussian noise and contrast). (\ref{fig:var_finetune_vs_finetune}): Pretraining and finetuning BERT medium with 10 different pretraining seeds and 10 different finetuning seeds per pretrained model leads to variation in accuracies on the Antonymy (yellow scatter points) and Numerical (blue scatter points) stress tests \cite{naik2018stress}.}
    \label{fig:display_variation}
\end{figure*}

First consider the intuitive layer specificity test (Section \ref{section:disagreement-layer}), which revealed that random initialization affects PWCCA more than large changes in layer depth. To justify why this is undesirable, we can turn to probing accuracy, which is strongly affected by layer depth, and only weakly affected by random seed (Figure \ref{fig:var_layer_exp}). This suggests a path forward: we can ground the layer test in the concrete differences in functionality captured by the probe.

More generally, we want metrics to be sensitive to changes that affect functionality, while ignoring those that don't. This motivates 
the following general procedure, given a distance metric $d$ and a functionality $f$ (which assigns a real number to a given representation):
\begin{enumerate}
	\item Collect a set $S$ of representations that differ along one or more axes of interest (e.g.~layer depth, random seed).
	\item Choose a reference representation $A \in S$. When $f$ is an accuracy metric, it is reasonable to choose $A = \arg\max_{A \in S} f(A)$.\footnote{Choosing the highest accuracy model as the reference makes it more likely that as accuracy changes, models are on average becoming more dissimilar. A low accuracy model may be on the ``periphery'' of model space, where it is dissimilar to models with high accuracy, but potentially even more dissimilar to other low accuracy models that make different mistakes.}
	\item For every representation $B \in S$:
	\begin{itemize}
		\item Compute $|f(A) - f(B)|$
		\item Compute $d(A, B)$
	\end{itemize}
	\item Report the rank correlation between $|f(A) - f(B)|$ and $d(A, B)$ (measured by Kendall's $\tau$ or Spearman $\rho$).
\end{enumerate}
The above procedure provides a \emph{quantitative} measure of how well the distance metric $d$ responds to the functionality $f$. 
For instance, in the layer specificity test, since depth affects probing accuracy strongly while random seed affects it only weakly, a dissimilarity measure with high rank correlation 
will be strongly responsive to layer depth and weakly responsive to seed; thus rank 
correlation quantitatively formalizes
the test from Section \ref{section:disagreement-layer}.   

Correlation metrics also capture properties 
that our intuition might miss.
For instance, 
Figure \ref{fig:var_layer_exp} shows that some variation in random seed actually does affect accuracy, and our procedure rewards metrics that pick up on this, 
while the intuitive sensitivity test would penalize them. 

Our procedure requires choosing a collection of models $S$; the crucial feature of $S$ is that it contains models with diverse behavior according to $f$. Different sets $S$, combined with a 
functional difference $f$, can be thought of as miniature ``benchmarks" that surface complementary perspectives on dissimilarity measures' 
responsiveness to that functional difference. In the rest of this section, we instantiate this quantitative benchmark for several choices of $f$ and $S$, starting with the layer and principal component tests from Section \ref{section:disagreement} and continuing on to several tests of OOD performance.

The overall results are summarized in Table~\ref{table:results}.
Note that for any single benchmark, 
we expect the correlation coefficients to be significantly lower than $1$, since the metric $D$ must capture all important axes of variation 
while $f$ measures only one type of functionality. A good metric is one that has consistently high correlation across many 
different functional measures.

\paragraph{Benchmark 1: Layer depth.} 

We turn the layer test into a benchmark for both text and images. For the text setting, we construct a set $S$ of $120$ representations by pretraining $10$ BERT base models with different initialization seeds and including each of the $12$ BERT layers as a representation. 
We separately consider two functionalities $f$: probing accuracy on QNLI \citep{wang2018glue} and SST-2 \citep{socher2013recursive}. To compute the rank correlation, 
we take the reference representation $A$ to be the representation with highest probing accuracy.
We compute the Kendall's $\tau$ and Spearman's $\rho$ rank correlations between the dissimilarities and the probing accuracy differences and report the results in Table \ref{table:results}. 

For the image setting, we similarly construct a set $S$ of $70$ representations by training $5$ ResNet-14 models with different initialization seeds and including each of the $14$ layers' representations. We also consider two functionalities $f$ for these vision models: probing accuracy on CIFAR-100 \cite{krizhevsky2009learning} and on SVHN \cite{netzer2011reading}, and compute rank correlations in the same way.

We find that PWCCA has lower rank correlations compared to CKA and Procrustes for both language probing tasks. This corroborates the intuitive specificity test (Section \ref{section:disagreement-layer}), 
suggesting that PWCCA registers too large of a dissimilarity across random initializations. For the vision tasks, CKA and Procrustes achieve similar rank correlations, while PWCCA cannot be computed because $n < d$.

% results table
\input{table_no_pval}

\paragraph{Benchmark 2: Principal component (PC) deletion.} 

We next quantify the PC deletion test from Section \ref{section:disagreement-pca},
by constructing a set $S$ of representations that vary in both random initialization and fraction of principal components deleted. We pretrain 10 BERT base models with different initializations, 
and for each pretrained model we obtain 14 different representations by deleting that representation's \(k\) smallest principal components, with \(k \in \{0, 100, 200, 300, 400, 500, 600, 650, 700, 725, 750, 758, 763, 767\}\). Thus \(S\) has \(10 \times 14 = 140\) elements. 
The representations themselves are the layer-$\ell$ activations, for $\ell \in \{8, 9, \ldots, 12\}$,\footnote{Earlier layers have near-chance accuracy on probing tasks, so we ignore them.} so there are $5$ different choices of $S$.
We use SST-2 probing accuracy as the functionality of interest $f$, and select the reference representation $A$ as the element in $S$ with highest accuracy. Rank correlation results are consistent across the 5 choices of $S$ (Appendix \ref{appendix:layer-wise-results}), so we report the average as a summary statistic in Table \ref{table:results}.

We find that PWCCA has the highest rank correlation between dissimilarity and probing accuracy, followed by Procrustes, and distantly followed by CKA. This corroborates the intuitive observations from
Section~\ref{section:disagreement-pca} that CKA is not sensitive to principal component deletion.

\subsection{Investigating variation in OOD performance across random seeds}

So far our benchmarks have been based on probing accuracy, which only measures in-distribution behavior (the train and test set of the probe are typically i.i.d.). 
In addition, the BERT models were always pretrained on language modeling but not finetuned for classification. To add diversity to our benchmarks, we next consider the out-of-distribution performance of language and vision models trained for classification tasks.
%of several collections of fine-tuned models.

\paragraph{Benchmark 3: Changing fine-tuning seeds.}
\citet{mccoy2020berts} show that a single pretrained BERT base model finetuned on MNLI with different random initializations will produce models with similar in-distribution performance, but widely variable performance on out-of-distribution data. We thus create a benchmark $S$ out of \citeauthor{mccoy2020berts}'s 100 released fine-tuned models, using OOD accuracy on the ``Lexical Heuristic (Non-entailment)" subset of the HANS dataset \citep{mccoy2019right} as our functionality $f$. This functionality is associated with the entire model, rather than an individual layer (in contrast to the probing functionality), but we consider one layer at a time to measure whether dissimilarities between representations at that layer correlate with $f$. This allows us to also localize whether certain layers are more predictive of $f$.

We construct 
$12$ different $S$ (one for each of the 12 layers of BERT base), taking the reference 
representation $A$ to be that of the 
highest accuracy model according to $f$. 
As before, we report each dissimilarity measure's rank correlation with $f$ in Table \ref{table:results}, averaged over the 12 runs. 

All three dissimilarity measures correlate with OOD accuracy, with Orthogonal Procrustes and PWCCA being more correlated than CKA. 
Since the representations in our benchmarks were computed on in-distribution MNLI data, this 
has the interesting implication that dissimilarity measures can detect OOD differences 
without access to OOD data. 
It also implies that random initialization leads to meaningful functional differences that 
are picked up by these measures, especially Procrustes and PWCCA. Contrast this with 
our intuitive specificity test in Section~\ref{section:disagreement-layer}, where all
sensitivity to random initialization was seen as a shortcoming. Our more quantitative benchmark 
here suggests that some of that sensitivity tracks true functionality.

To check that the differences in rank correlation for Procrustes, PWCCA, and CKA are statistically significant, we compute bootstrap estimates of their 95\% confidence intervals. With 2000 bootstrapped samples,  we find statistically significant differences between all pairs of measures for most choices of layer depth $S$, so we conclude PWCCA > Orthogonal Procrustes > CKA (the full results are in Appendix \ref{appendix:bootstrap}). We do not apply this procedure for the previous two 
benchmarks, because the different models have correlated randomness and so any $p$-value based on independence assumptions would be invalid.

\paragraph{Benchmark 4: Challenge sets: Changing pretraining and fine-tuning seeds.} We also construct benchmarks using models trained from scratch with different random seeds (for language, this is pretraining and fine-tuning, and for vision, this is standard training). For language, we construct benchmarks from a collection of 100 BERT medium models, trained with all combinations of 10 pretraining and 10 fine-tuning seeds. 
The models are fine-tuned on MNLI, and we consider two different functionalities of interest $f$: accuracy on the OOD Antonymy stress test and on the OOD Numerical stress test \citep{naik2018stress}, which both show significant variation in accuracy 
across models (see Figure \ref{fig:var_finetune_vs_finetune}). We obtain $8$ different sets $S$ (one for each of the 8 layer depths in BERT medium), again taking $A$ to be the representation of the highest-accuracy model according to $f$. 
Rank correlations for each dissimilarity measure are averaged over the 8 runs and reported in Table \ref{table:results}. 

For vision, we construct benchmarks from a collection of 100 ResNet-14 models, trained with different random seeds on CIFAR-10. 
We consider 19 different functionalities of interest---the 19 types of corruptions in the CIFAR-10C dataset \cite{hendrycks2019robustness}, which show significant variation in accuracy across models (see Figure \ref{fig:var_cifar10c}). 
We obtain $14$ different sets $S$ (one for each of the 14 layers), taking $A$ to be the representation of the highest-accuracy model according to $f$. 
Rank correlations for each dissimilarity measure are averaged over the 14 runs and over the 19 corruption types and reported in Table \ref{table:results}. Results for each of the 19 corruptions individually can be found in Appendix \ref{appendix:cifar10c}..

None of the dissimilarity measures show a large rank correlation for either the language or vision tasks, and for the Numerical stress test, at most layers, the associated $p$-values (assuming independence) are non-significant at the 0.05 level (see Appendix \ref{appendix:layer-wise-results}).
\footnote{See Appendix \ref{appendix:layer-wise-results} for $p$-values as produced by sci-kit learn. Strictly speaking, the $p$-values are invalid because they assume independence, but the pretraining seed induces correlations. However, correctly accounting for these would tend to make the $p$-values larger, thus preserving our conclusion of non-significance
.} Thus we conclude that all measures fail to be sensitive to OOD accuracy in these settings. 
One reason for this could be that there is less variation in the OOD accuracies compared to the previous experiment with the HANS dataset (there accuracies varied from 0 to nearly 60\%). Another reason could be that 
it is harder to correctly account for both pretraining and 
fine-tuning variation at the same time.
Either way, we hope 
that future dissimilarity measures can improve upon these results, and we present this benchmark as a challenge task to motivate progress.

%% file: table_no_pval.tex
\begin{table}[t]
\centering
	\caption{\textbf{Summary of rank correlation results.} For Benchmarks \#1-3 in both language and vision, all dissimilarity measures successfully achieve significant positive rank correlation with the functionality of interest--both CKA and PWCCA dominate certain benchmarks and fall behind on others, while Procrustes is more consistent and often close to the leader. 
Benchmark \#4 is more challenging, and no dissimilarity measure achieves a high correlation. The vision experiments do not have results for PWCCA because $n < d$. } 
	\label{table:results}
	\begin{tabular}{@{}cccccccccc@{}}
	\toprule
	
\multirow{2}{*}{\#} 
	&\multirow{2}{*}{Perturbation}                                                             
  & \multirow{2}{*}[-0.2em]{\begin{tabular}[c]{@{}c@{}}Subtask\\ Size\end{tabular}}

	& \multirow{2}{*}{Functionality}                                  
	& \multicolumn{2}{c}{Procrustes}                                                                                                          
	& \multicolumn{2}{c}{CKA}                                                                                             
	& \multicolumn{2}{c}{PWCCA}                                                                                                             
	
	\\ \cmidrule(l){5-10} 
																							  
	& &  & $\rho$  & $\tau$    & $\rho$   & $\tau$    & $\rho$   & $\tau$                                                        
	 \\ \midrule
	 \\
	 \multicolumn{10}{c}{Modality: Language}\\
	 \midrule
	 	\multirow{2}{*}{1}&\multirow{2}{*}{\begin{tabular}[l]{@{}l@{}}Pretraining seed, 
	\\ layer depth\end{tabular}} & 120 & Probe: QNLI & 0.862   & 0.670    & \textbf{0.876}   & \textbf{0.685}   & 0.763  & 0.564  
	\\ \cmidrule(l){3-10} 
	&& 120 & Probe: SST-2  & 0.890  & 0.707 & \textbf{0.905}   & \textbf{0.732}  & 0.829   & 0.637                                                              
	\\ \midrule
	2&\begin{tabular}[l]{@{}l@{}}Pretraining seed, \\ PC deletion\end{tabular}     
	 & $140 \times 5$ & Probe: SST-2
	& 0.860  & 0.677  & 0.751   & 0.564   & \textbf{0.870}  & \textbf{0.690} 
	 \\ \midrule
	3&Finetuning seed & $100 \times 12$
	& \begin{tabular}[c]{@{}c@{}}OOD: HANS\\ Lexical non-entailed\end{tabular} & 0.551 & 0.398 & 0.462 & 0.329 & \textbf{0.568} & \textbf{0.412} \\
	%& \begin{tabular}[c]{@{}c@{}}HANS: Lexical \\ (non-entailed)\end{tabular} 
	%& \begin{tabular}[c]{@{}c@{}}0.551 \end{tabular} & \begin{tabular}[c]{@{}c@{}}0.398 \end{tabular} & \begin{tabular}[c]{@{}c@{}}0.462\end{tabular} & \begin{tabular}[c]{@{}c@{}}0.329\end{tabular} & \textbf{\begin{tabular}[c]{@{}c@{}}0.568\end{tabular}} & \textbf{\begin{tabular}[c]{@{}c@{}}0.412 \end{tabular}} \\ 
	
	\midrule
	
	\multirow{3}{*}{4}&\multirow{3}{*}{\begin{tabular}[l]{@{}l@{}}Pretraining and \\ finetuning seeds\end{tabular}} & $100 \times 8$ & \begin{tabular}[c]{@{}c@{}}OOD: Antonymy\\ stress test\end{tabular} & \textbf{0.243} & \textbf{0.178} & 0.227 & 0.160  & 0.204  & 0.152 \\ 
	
	\cmidrule(l){3-10} &&	$100 \times 8$	  & \begin{tabular}[c]{@{}c@{}}OOD: Numerical\\ stress test\end{tabular} & 0.071 & 0.049 & \textbf{0.122}  & \textbf{0.084}  & 0.031  & 0.023 \\

	\midrule

&Total (language) & $3740$ & Average & \textbf{0.580} & \textbf{0.447} & 0.557 & 0.426 & 0.544 & 0.413\\
\midrule
\\
	\multicolumn{10}{c}{Modality: Vision}\\
	 \midrule
	 \multirow{2}{*}{1}& \multirow{2}{*}{\begin{tabular}[l]{@{}l@{}}Training seed, 
	\\ layer depth\end{tabular}} & 70 & Probe: CIFAR-100 & 0.485   & \textbf{0.376}   & \textbf{0.507}   & 0.359   & -  & -  
	\\ \cmidrule(l){3-10} 
	&& 70 & Probe: SVHN  & 0.363  & \textbf{0.272} & \textbf{0.372}   & 0.255  & -   & -                                                              
	\\ 
	 
	 \midrule
	 4& Training seed & $1900 \times 14$ & OOD: CIFAR-10C & \textbf{0.060} & \textbf{0.057} & 0.041 & 0.038 & - & - \\
	 \midrule

&Total (vision) & $26740$ & Average & 0.303 & \textbf{0.235} & \textbf{0.307} & 0.217 & - & - 
	\\ 
	\bottomrule
	\end{tabular}
\end{table}

%% file: discussion.tex
In this work we proposed a quantitative measure for 
evaluating similarity metrics, based on the rank correlation with 
functional behavior. Using this, we generated tasks motivated by 
sensitivity to deleting important directions, specificity to 
random initialization, and sensitivity to out-of-distribution 
performance. Popular existing metrics such as CKA and CCA often 
performed poorly on these tasks, sometimes in striking ways. 
Meanwhile, the classical Orthogonal Procrustes transform attained 
consistently good performance.

Given the success of Orthogonal Procrustes, it is worth reflecting 
on how it differs from the other metrics and why it might perform 
well. To do so, we consider a simplified case where 
$A$ and $B$ have the same singular vectors but different singular 
values. Thus without loss of generality $A = \Lambda_1$ and 
$B = \Lambda_2$, where the $\Lambda_i$ are both diagonal.
In this case, the Orthogonal Procrustes distance reduces to 
$\|\Lambda_1 - \Lambda_2\|_F^2$, or the sum of the squared distances 
between the singular values. We will see that both CCA and CKA 
reduce to less reasonable formulae in this case.

\emph{Orthogonal Procrustes vs.~CCA.} All three metrics derived 
from CCA assign \emph{zero} distance even when the (non-zero) 
singular values are arbitrarily different. This is because CCA 
correlation coefficients are invariant to all invertible linear 
transformations.  
This invariance property may help explain why CCA metrics generally find layers within the same network to be much more similar than networks trained with different randomness. 
Random initialization introduces noise, 
particularly in unimportant principal components, while representations within the same network more easily preserve these components, and CCA may place too much weight on their associated correlation coefficients.

\emph{Orthogonal Procrustes vs.~CKA.} 
In contrast to the squared distance of Orthogonal Procrustes, 
CKA actually reduces to a quartic function based on the dot 
products between the \emph{squared} entries of 
$\Lambda_1$ and $\Lambda_2$. As a consequence, 
CKA is dominated by representations' largest singular values, leaving it insensitive to meaningful differences in smaller singular values 
as illustrated in Figure \ref{fig:disagree-pca}. This lack of 
sensitivity to moderate-sized differences may help 
explain why CKA fails to track out-of-distribution error effectively.

In addition to helping understand similarity measures, our benchmarks 
pinpoint directions for improvement.
No method was sensitive to accuracy on the Numerical stress test 
in our challenge set, possibly due to a lower signal-to-noise ratio. Since Orthogonal Procrustes performed well on most of our tasks, it could be a promising foundation for a new measure, and recent work shows how to regularize Orthogonal Procrustes to handle high noise \citep{pumir2021generalized}. Perhaps similar techniques could be adapted here. 
  
An alternative to our benchmarking approach is to directly define two representations' dissimilarity as their difference in a functional behavior of interest. \citet{feng2020transferred} take this approach, defining dissimilarity as difference in accuracy on a handful of probing tasks. One drawback of this approach is that a small set of probes may not capture all the differences in representations, so it is useful to base dissimilarity measures on representations’ intrinsic properties. Intrinsically defined dissimilarities also have the potential to highlight new functional behaviors, as we found that representations with similar in-distribution probing accuracy often have highly variable OOD accuracy.

A limitation of our work is that we only consider a handful of model variations and functional behaviors, and restricting our attention to these settings could overlook other important considerations. 
To address this, we envision a paradigm in which a rich tapestry of benchmarks 
are used to ground and validate neural network interpretations.  
Other axes of variation in models could include training on more or fewer examples, training on shuffled labels vs.~real labels, training from specifically chosen initializations \citep{frankle2018lottery}, and using different architectures. 
Other functional behaviors to examine could include modularity and meta-learning capabilities. 
Benchmarks could also be applied to other interpretability tools beyond dissimilarity. For example, sensitivity to 
deleting principal components could provide an additional 
sanity check for saliency maps and other visualization tools 
\citep{adebayo2018sanity}.

More broadly, many interpretability tools are designed as \emph{audits} of models, although it is often unclear what characteristics of the models are consistently audited. 
We position this work as a \emph{counter-audit}, where by collecting models that differ in functional behavior, we can assess whether the interpretability tools CKA, PWCCA, etc., accurately reflect the behavioral differences. 
Many other types of counter-audits may be designed to assess other interpretability tools. 
For example, models that have backdoors built into them to misclassify certain inputs provide counter-audits for interpretability tools that explain model predictions--these explanations should reflect any backdoors present \citep{li2020deep,chen2017targeted,wang2019neural,kurita2020weight}. %\js{cite a few other representative papers}. 
We are hopeful that more comprehensive checks on interpretability tools will provide deeper understanding of neural networks, and more reliable models.

%% file: appendix.tex
\section*{Appendix}

\section{Training details} \label{appendix:finetuning}
\subsection{BERT finetuning details}
We fine-tuned models from \citet{zhong2021larger} and the original BERT models  from \citet{devlin2018bert} on three tasks -- Quora Question Pairs (QQP)\footnote{https://www.quora.com/q/quoradata/First-Quora-Dataset-Release-Question-Pairs}, Multi-Genre Natural Language Inference (MNLI; \citet{N18-1101}), and the Stanford Sentiment Treebank (SST-2; \citet{socher2013recursive}), and show each model's accuracy on these tasks in Table \ref{table:compare-to-orig}. 
Our models generally have comparable accuracy.

As in \citet{turc2019well}, we finetune for 4 epochs for each dataset. For each task and model size, we tune hyperparameters in the
following way: we first randomly split our new
training set into 80\% and 20\%; then we finetune on
the 80\% split with all 9 combination of batch size
[16, 32, 64] and learning rate [1e-4, 5e-5, 3e-5],
and choose the combination that leads to the best
average accuracy on the remaining 20\%. Finetuning these models for all three tasks requires around 500 hours.

\begin{table}[h]
    \centering
    \caption{Comparing accuracy of our pretrained model (superscript $^{ours}$) to the original release by \citet{devlin2018bert} and \citet{turc2019well} (superscript $^{orig}$) on a variety of fine-tuned tasks.}
    \begin{tabular}{lrrr}
    \toprule
     &   QQP &  MNLI &  SST-2 \\
    \hline
    %$\text{Mini}^{\text{ orig}}$   & 88.2\% & 74.6\% &  92.8\% \\
    %$\text{Mini}^{\text{ ours}}$   & 87.3\% & 74.3\% &  92.8\% \\
    %\hline
    %$\text{Small}^{\text{ orig}}$  & 89.1\% & 77.3\% &  93.9\% \\
    %$\text{Small}^{\text{ ours}}$  & 88.7\% & 76.7\% &  93.9\% \\
    %\hline
    $\text{BERT medium}^{\text{ orig}}$ & 89.8\% & 79.6\% &  94.2\% \\
    $\text{BERT medium}^{\text{ ours}}$ & 89.5\% & 78.9\% &  94.2\% \\
    \hline
    $\text{BERT base}^{\text{ orig}}$   & 90.8\% & 83.8\% &  95.0\% \\
    $\text{BERT base}^{\text{ ours}}$   & 90.6\% & 81.2\% &  94.6\% \\
    %\hline
    %$\text{Large}^{\text{ orig}}$  & 91.3\% & 86.8\% &  95.2\% \\
    %$\text{Large}^{\text{ ours}}$  & 91.0\% & 83.8\% &  94.8\% \\
    \bottomrule
    \end{tabular}
    
    \label{table:compare-to-orig}
\end{table}

\subsection{ResNet training details}
We trained ResNet-14 models on CIFAR-10 training data with the following hyperparameters:
\begin{itemize}
	\item learning rate: 0.1
	\item epochs: 100
	\item learning rate decay: 0.1 at epoch 50 and epoch 75
	\item batch size: 128
\end{itemize}
The 100 models we trained have an average accuracy on the CIFAR-10 test set of 90.2\%, with standard deviation 0.2\%. Training these models requires around 20 hours.

\section{Licenses}\label{appendix:licenses}

The source code for BERT models available at \url{https://github.com/google-research/bert} is licensed under the Apache License 2.0.

The model weights for the 100 BERT base models provided by \citet{mccoy2020berts} are licensed under the Creative Commons Attribution 4.0 International license, and their source code is licensed under the MIT license (\url{https://github.com/tommccoy1/hans/blob/master/LICENSE.md}).

\section{Layer-wise results}\label{appendix:layer-wise-results}
\input{layer_tables}

\section{CIFAR-10C subtask-wise results}\label{appendix:cifar10c}
\input{cifar10c_all_table.tex}

\section{Bootstrap significance testing for changing fine-tuning seeds}\label{appendix:bootstrap}

To assess whether the differences between rank correlations are statistically significant in the experiments varying finetuning seed and comparing functional behavior on the OOD HANS dataset, we conduct bootstrap resampling.
Concretely, for every pair of metrics and every layer depth, we do the following:
\begin{itemize}
	\item Sample 100 models with replacement, and collect their representations at the specified layer depth
    \item Let the reference $A$ be the representation corresponding to the sampled model with maximum accuracy at that depth
    \item Compute the dissimilarities between $A$ and the 100 sampled representations
    \item Compute the Kendall's $\tau$ and Spearman's $\rho$ rank correlations for Orthogonal Procrustes, CKA, and PWCCA
    \item Record $\rho$(Procrustes) - $\rho$(CKA), $\rho$(PWCCA) - $\rho$(CKA), and $\rho$(PWCCA) - $\rho$(Procrustes), and the same pairwise differences for Kendall's $\tau$.
    \item Repeat the above 2000 times
\end{itemize}

This gives us bootstrap distributions for the differences in rank correlations, and we may compute the 95\% confidence intervals for these distributions. When the confidence interval does not overlap with 0, we conclude that the difference in rank correlation is statistically significant. The figures below show the results for each layer. We see that in the deeper layers of the network (layers 8-12), PWCCA has statistically significantly higher rank correlation than Orthogonal Procrustes, which in turn has statistically significantly higher rank correlation than CKA. In earlier layers, results are sometimes statistically significant, but not always. 

\begin{figure*}[h]
    \centering
    \caption{Bootstrap comparison of $\rho$ between metrics, layers 1-4}
    \includegraphics[width=13.5cm]{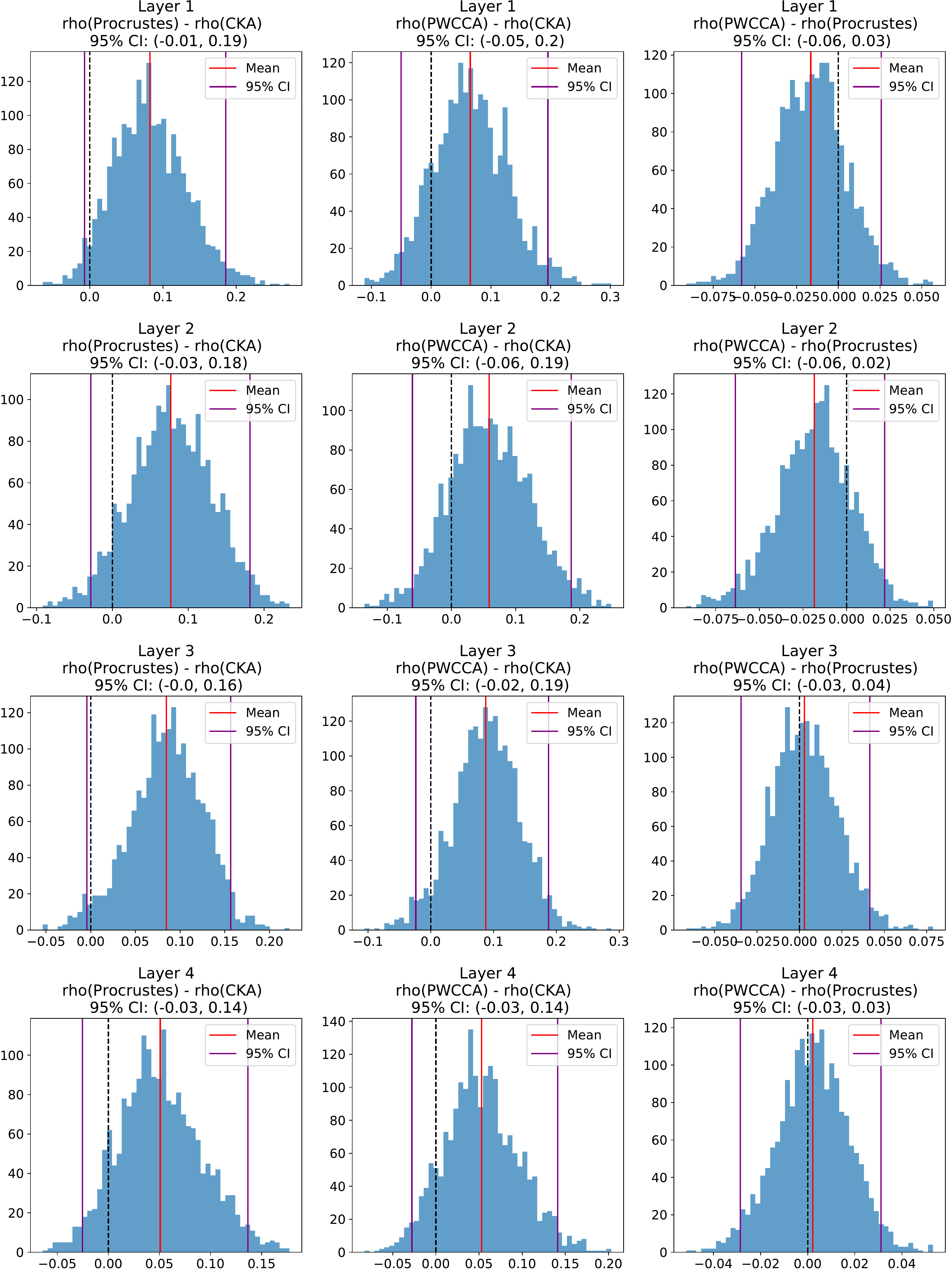}
    \label{fig:histograms_rho04}
\end{figure*}
\begin{figure*}[h]
    \centering
    \caption{Bootstrap comparison of $\rho$ between metrics, layers 5-8}
    \includegraphics[width=13.5cm]{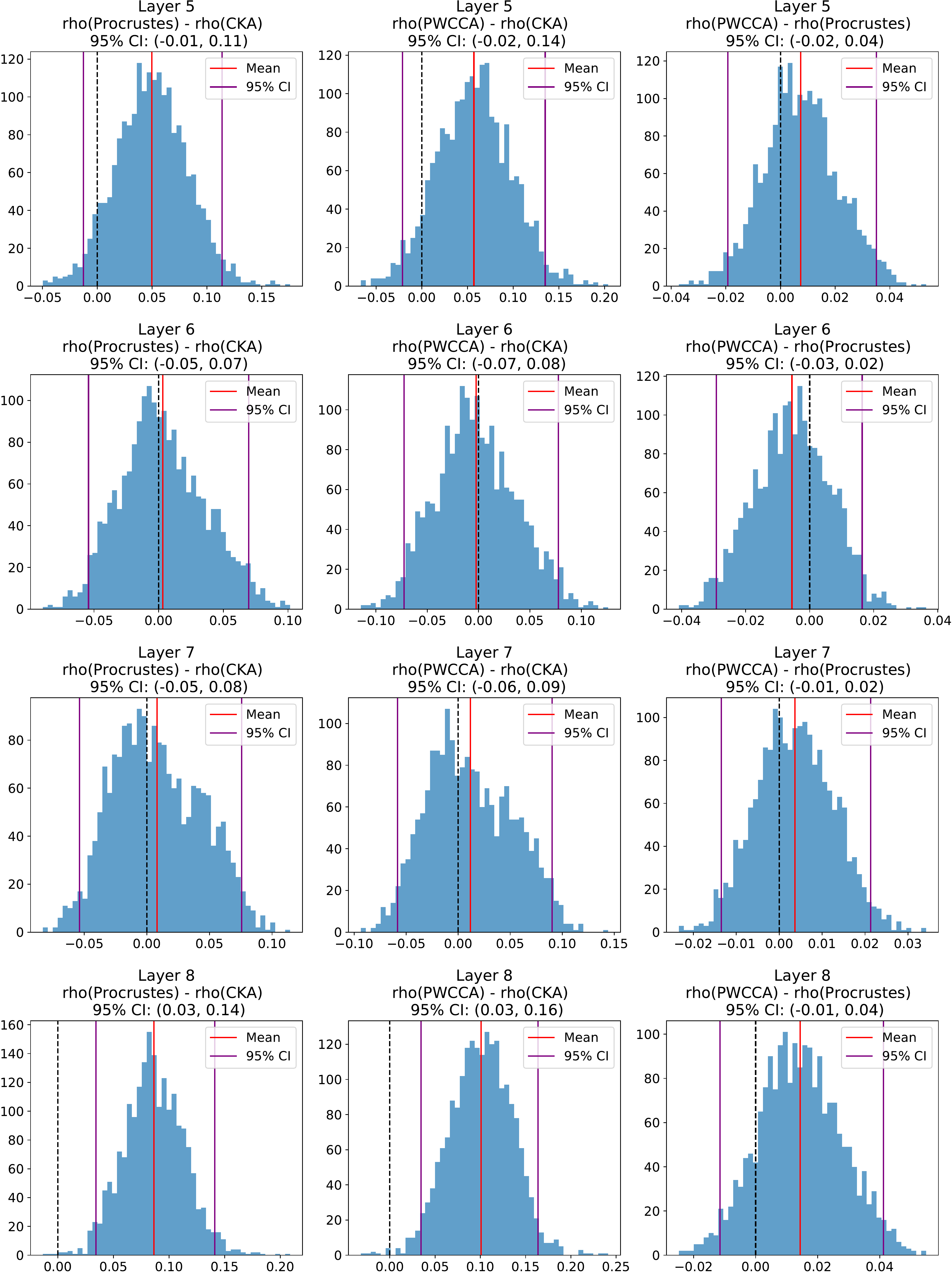}
    \label{fig:histograms_rho48}
\end{figure*}
\begin{figure*}[h]
    \centering
    \caption{Bootstrap comparison of $\rho$ between metrics, layers 9-12}
    \includegraphics[width=13.5cm]{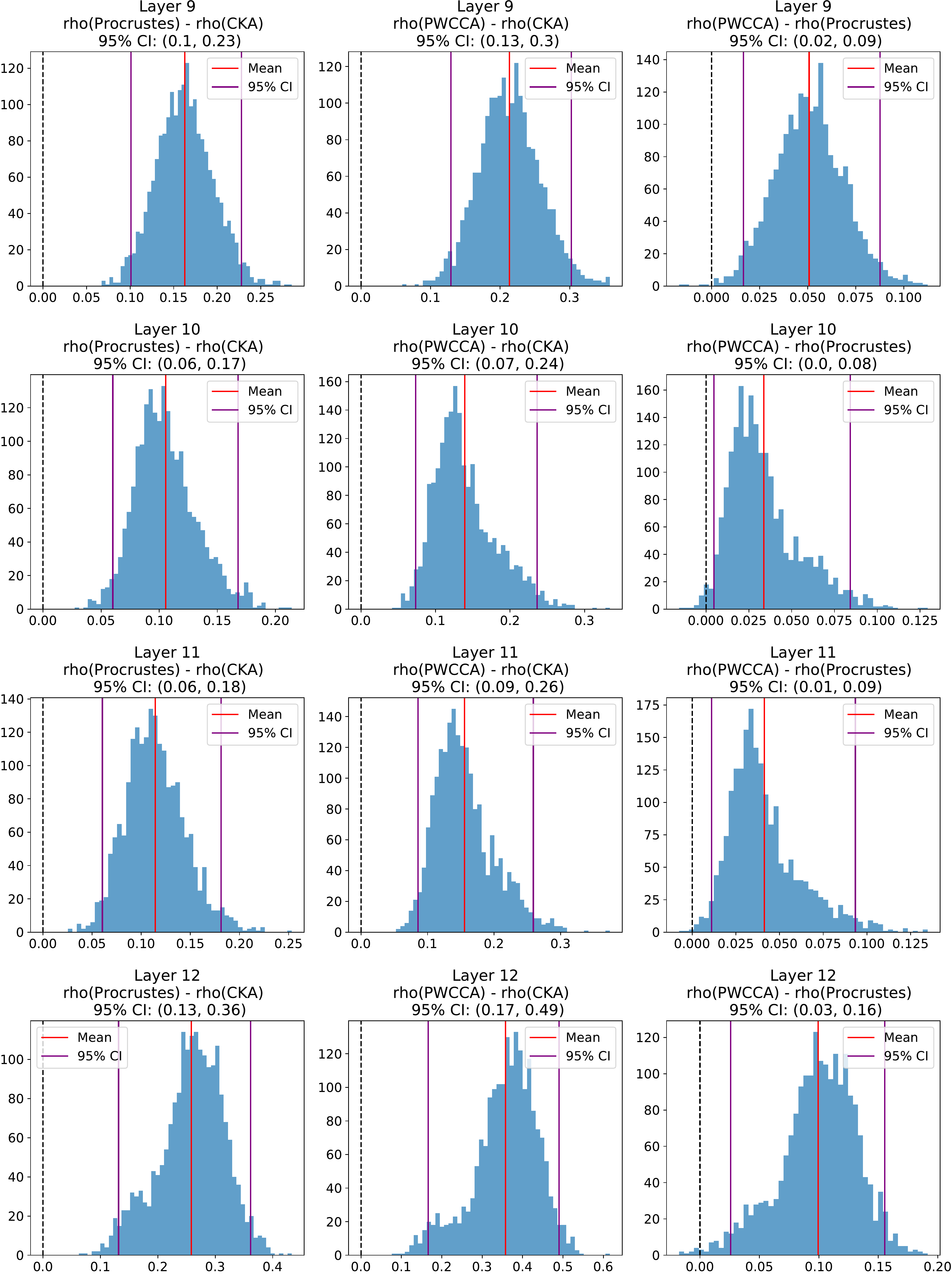}
    \label{fig:histograms_rho812}
\end{figure*}

\begin{figure*}[h]
    \centering
    \caption{Bootstrap comparison of $\tau$ between metrics, layers 1-4}
    \includegraphics[width=13.5cm]{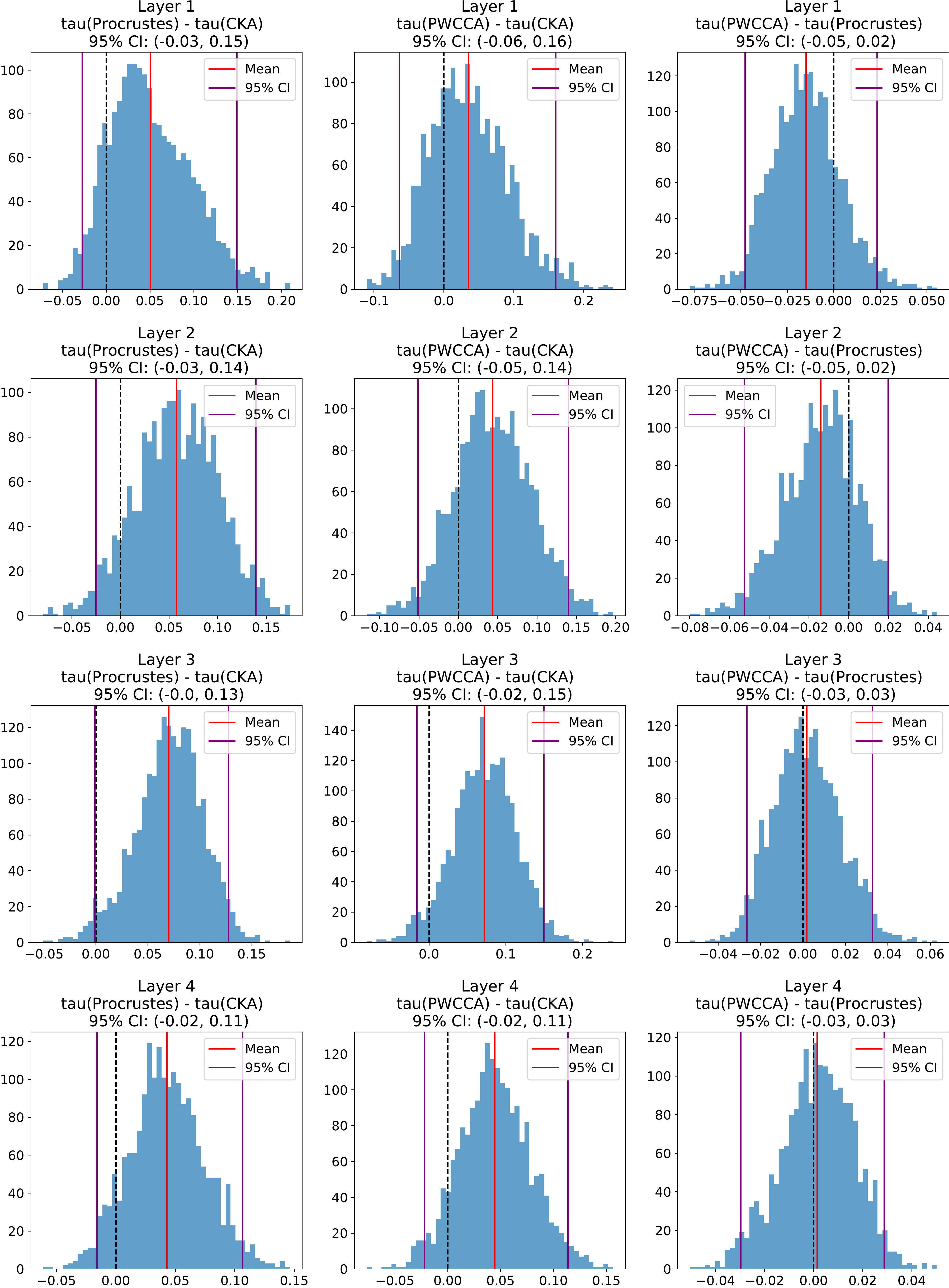}
    \label{fig:histograms_tau04}
\end{figure*}

\begin{figure*}[h]
    \centering
    \caption{Bootstrap comparison of $\tau$ between metrics, layers 5-8}
    \includegraphics[width=13.5cm]{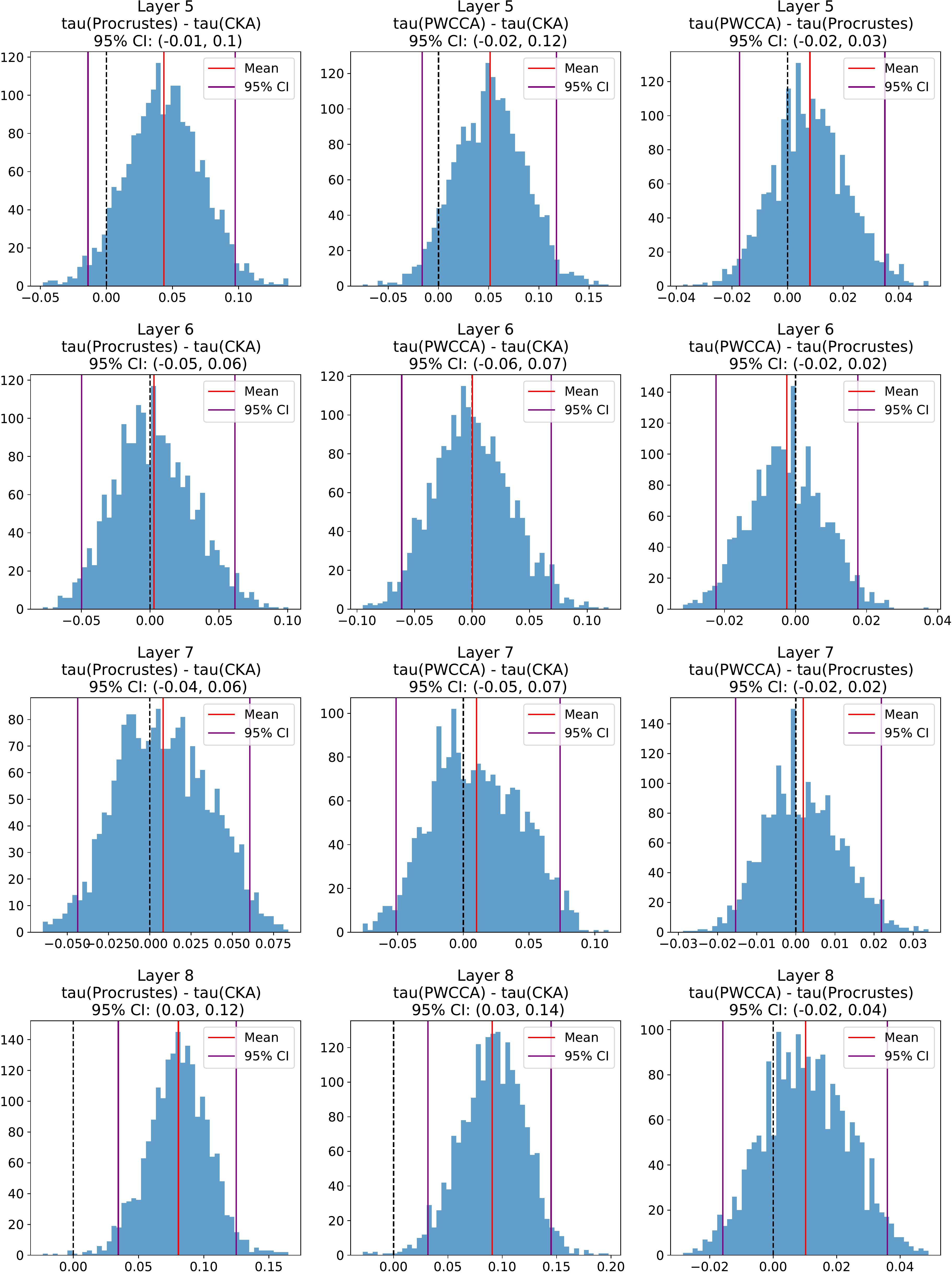}
    \label{fig:histograms_tau48}
\end{figure*}

\begin{figure*}[h]
    \centering
    \caption{Bootstrap comparison of $\tau$ between metrics, layers 9-12}
    \includegraphics[width=13.5cm]{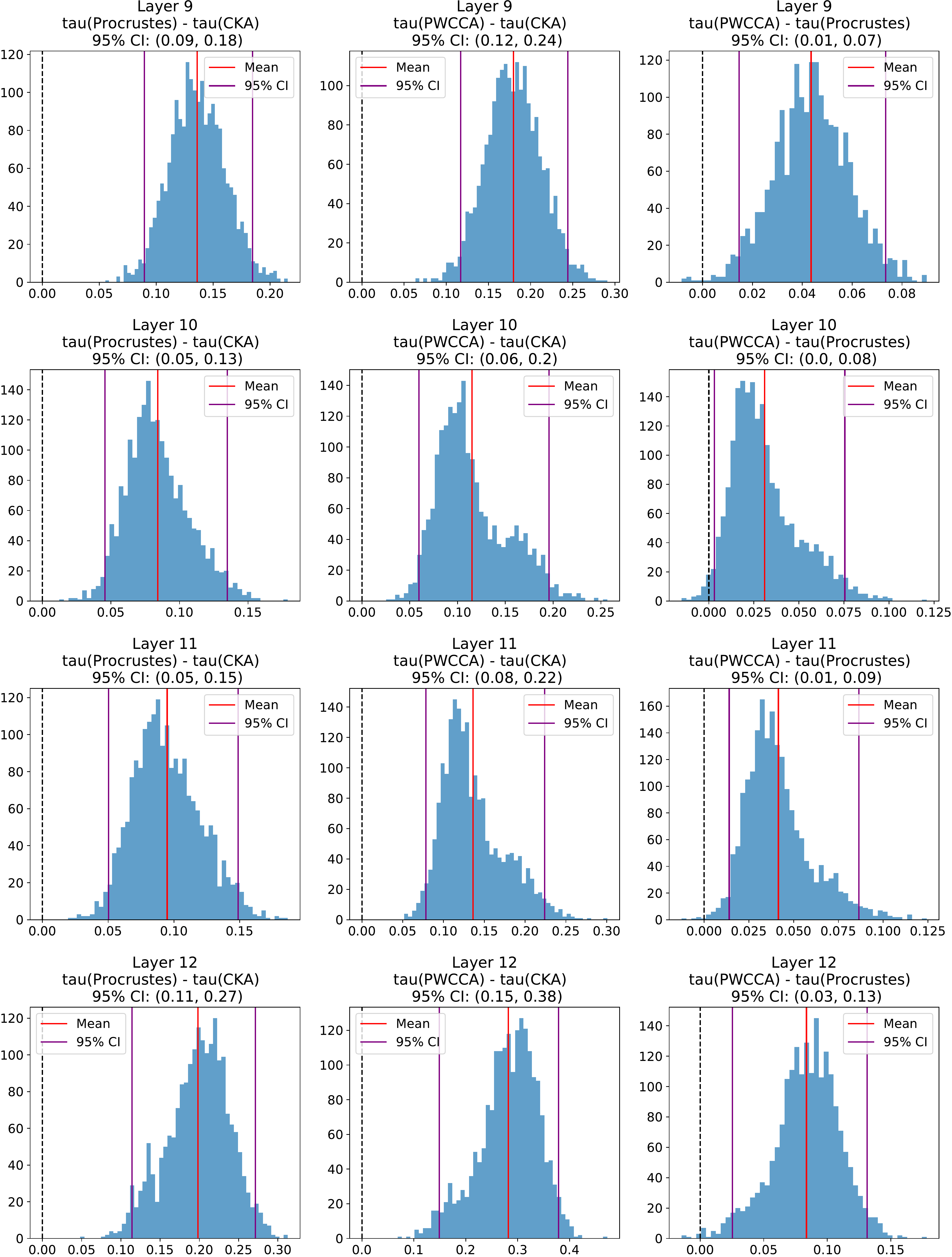}
    \label{fig:histograms_tau812}
\end{figure*}

\clearpage

%% file: layer_tables.tex
Some of the results presented in Table \ref{table:results} were averaged over multiple layers, since rankings between dissimilarity measures were consistent across different layers. Rank correlation scores are higher across all measures for certain layers, however, so we include layer-by-layer results here for completeness. We also include scores for $\bar \rho_{\text{CCA}}$ and $R^2_{\text{CCA}}$ here, and note that they are often similar to PWCCA, and generally dominated by other measures. We expand each row of Table \ref{table:results} into a subsection of its own. We also include p-values as reported by sci-kit learn, although we note that because random seeds are shared among some representations, these p-values are all inflated, with the exception of those for the experiment perturbing only fine-tuning seed, and assessing functionality through HANS (\ref{appendix:feathertable}). The invalid p-values may all be thought of as upper-bounds for the significance of the rank correlation results.

\subsection{Perturbation: pretraining seed and layer depth}

%We find for this task that the first and last layers of BERT base achieve high rank correlations, across all dissimilarity measures, while the middle layers achieve significantly worse rank correlations. 
Tables \ref{table:layer-qnli-rho} and \ref{table:layer-qnli-tau} show the full results (including p-values and all 5 dissimilarity measures) using the QNLI probe as the functionality of interest, for Spearman $\rho$ and Kendall's $\tau$, respectively. Table \ref{table:layer-sst2-rho} and \ref{table:layer-sst2-tau} present results for the probing task SST-2 as the functionality of interest.

% QNLI rho
\begin{table}[h]
\centering
	\caption{Spearman $\rho$ results for perturbing pretraining seed and layer depth, and assessing functionality through the QNLI probe}
	\label{table:layer-qnli-rho}
	\begin{tabular}{cccccc}
		\toprule
		Layer & Procrustes & CKA & PWCCA & $\bar \rho_{\text{CCA}}$ & $R^2_{\text{CCA}}$ \\
		\midrule
			%1 & 0.821 (8.3E-31) & 0.832 (2.9E-32) & 0.787 (8.0E-27) & 0.823 (1.0E+00) & 0.823 (1.0E+00) \\
%2 & 0.860 (1.6E-36) & 0.851 (4.7E-35) & 0.850 (6.7E-35) & 0.865 (1.0E+00) & 0.865 (1.0E+00) \\
%3 & 0.867 (8.3E-38) & 0.853 (2.3E-35) & 0.855 (9.5E-36) & 0.868 (1.0E+00) & 0.866 (1.0E+00) \\
%4 & 0.420 (8.8E-07) & 0.485 (9.9E-09) & 0.316 (2.2E-04) & 0.321 (1.0E+00) & 0.311 (1.0E+00) \\
%5 & 0.512 (1.1E-09) & 0.560 (1.4E-11) & 0.421 (8.7E-07) & 0.441 (1.0E+00) & 0.440 (1.0E+00) \\
%6 & -0.118 (9.0E-01) & 0.023 (4.0E-01) & -0.187 (9.8E-01) & -0.264 (1.8E-03) & -0.286 (7.8E-04) \\
%7 & 0.063 (2.5E-01) & 0.225 (6.7E-03) & -0.076 (8.0E-01) & -0.076 (2.0E-01) & -0.111 (1.1E-01) \\
%8 & 0.402 (2.7E-06) & 0.448 (1.5E-07) & 0.252 (2.8E-03) & 0.334 (1.0E+00) & 0.285 (1.0E+00) \\
%9 & 0.614 (4.5E-14) & 0.699 (3.3E-19) & 0.507 (1.7E-09) & 0.596 (1.0E+00) & 0.577 (1.0E+00) \\
%10 & 0.755 (1.0E-23) & 0.832 (2.6E-32) & 0.656 (2.0E-16) & 0.742 (1.0E+00) & 0.736 (1.0E+00) \\
%11 & 0.835 (1.2E-32) & 0.857 (4.5E-36) & 0.746 (7.8E-23) & 0.805 (1.0E+00) & 0.803 (1.0E+00) \\
12 & 0.862 (6.5E-37) & 0.876 (1.6E-39) & 0.763 (2.2E-24) & 0.849 (1.0E+00) & 0.846 (1.0E+00) \\
		\bottomrule
	\end{tabular}
\end{table}

% QNLI tau
\begin{table}[h]
\centering
	\caption{Kendall's $\tau$ results for perturbing pretraining seed and layer depth, and assessing functionality through the QNLI probe}
	\label{table:layer-qnli-tau}
	\begin{tabular}{cccccc}
		\toprule
		Layer & Procrustes & CKA & PWCCA & $\bar \rho_{\text{CCA}}$ & $R^2_{\text{CCA}}$ \\
		\midrule
%			1 & 0.608 (3.4E-23) & 0.618 (7.2E-24) & 0.574 (7.8E-21) & 0.611 (1.0E+00) & 0.610 (1.0E+00) \\
%2 & 0.652 (2.4E-26) & 0.644 (8.9E-26) & 0.645 (7.7E-26) & 0.659 (1.0E+00) & 0.658 (1.0E+00) \\
%3 & 0.668 (1.4E-27) & 0.653 (1.8E-26) & 0.657 (1.0E-26) & 0.666 (1.0E+00) & 0.664 (1.0E+00) \\
%4 & 0.306 (3.6E-07) & 0.357 (3.7E-09) & 0.218 (2.1E-04) & 0.216 (1.0E+00) & 0.207 (1.0E+00) \\
%5 & 0.373 (7.8E-10) & 0.401 (4.2E-11) & 0.306 (3.7E-07) & 0.317 (1.0E+00) & 0.312 (1.0E+00) \\
%6 & -0.063 (8.5E-01) & 0.038 (2.7E-01) & -0.111 (9.6E-01) & -0.155 (5.9E-03) & -0.171 (2.8E-03) \\
%7 & 0.062 (1.6E-01) & 0.160 (4.8E-03) & -0.027 (6.7E-01) & -0.014 (4.1E-01) & -0.037 (2.7E-01) \\
%8 & 0.297 (7.9E-07) & 0.314 (1.8E-07) & 0.190 (1.1E-03) & 0.256 (1.0E+00) & 0.222 (1.0E+00) \\
%9 & 0.453 (1.1E-13) & 0.511 (6.8E-17) & 0.369 (1.1E-09) & 0.442 (1.0E+00) & 0.425 (1.0E+00) \\
%10 & 0.552 (2.0E-19) & 0.631 (8.6E-25) & 0.467 (1.9E-14) & 0.544 (1.0E+00) & 0.536 (1.0E+00) \\
%11 & 0.644 (8.9E-26) & 0.663 (3.6E-27) & 0.564 (3.3E-20) & 0.623 (1.0E+00) & 0.622 (1.0E+00) \\
12 & 0.670 (1.1E-27) & 0.685 (7.4E-29) & 0.564 (3.2E-20) & 0.652 (1.0E+00) & 0.647 (1.0E+00) \\
		\bottomrule
	\end{tabular}
\end{table}

% SST-2 rho
\begin{table}[h]
\centering
	\caption{Spearman $\rho$ results for perturbing pretraining seed and layer depth, and assessing functionality through the SST-2 probe}
	\label{table:layer-sst2-rho}
	\begin{tabular}{cccccc}
		\toprule
		Layer & Procrustes & CKA & PWCCA & $\bar \rho_{\text{CCA}}$ & $R^2_{\text{CCA}}$ \\
		\midrule
%			1 & 0.901 (6.4E-45) & 0.891 (1.7E-42) & 0.898 (3.6E-44) & 0.901 (1.0E+00) & 0.900 (1.0E+00) \\
%2 & 0.744 (1.1E-22) & 0.736 (5.5E-22) & 0.744 (1.0E-22) & 0.745 (1.0E+00) & 0.745 (1.0E+00) \\
%3 & 0.828 (1.1E-31) & 0.831 (3.3E-32) & 0.795 (9.9E-28) & 0.802 (1.0E+00) & 0.802 (1.0E+00) \\
%4 & 0.663 (8.0E-17) & 0.656 (2.1E-16) & 0.620 (2.3E-14) & 0.570 (1.0E+00) & 0.558 (1.0E+00) \\
%5 & -0.060 (7.4E-01) & 0.022 (4.1E-01) & -0.142 (9.4E-01) & -0.224 (6.9E-03) & -0.247 (3.3E-03) \\
%6 & 0.386 (6.7E-06) & 0.533 (1.9E-10) & 0.286 (7.7E-04) & 0.212 (9.9E-01) & 0.190 (9.8E-01) \\
%7 & 0.271 (1.4E-03) & 0.457 (7.7E-08) & 0.094 (1.5E-01) & 0.112 (8.9E-01) & 0.068 (7.7E-01) \\
%8 & 0.482 (1.3E-08) & 0.585 (1.2E-12) & 0.424 (6.8E-07) & 0.481 (1.0E+00) & 0.430 (1.0E+00) \\
%9 & 0.745 (8.5E-23) & 0.810 (2.2E-29) & 0.719 (1.2E-20) & 0.735 (1.0E+00) & 0.719 (1.0E+00) \\
%10 & 0.795 (1.0E-27) & 0.847 (1.6E-34) & 0.661 (1.1E-16) & 0.761 (1.0E+00) & 0.757 (1.0E+00) \\
%11 & 0.843 (7.2E-34) & 0.880 (2.7E-40) & 0.744 (9.8E-23) & 0.800 (1.0E+00) & 0.797 (1.0E+00) \\
12 & 0.890 (2.7E-42) & 0.905 (5.3E-46) & 0.829 (7.7E-32) & 0.857 (1.0E+00) & 0.854 (1.0E+00) \\
		\bottomrule
	\end{tabular}
\end{table}

% SST-2 tau
\begin{table}[!h]
\centering
	\caption{Kendall's $\tau$ results for perturbing pretraining seed and layer depth, and assessing functionality through the SST-2 probe}
	\label{table:layer-sst2-tau}
	\begin{tabular}{cccccc}
		\toprule
		Layer & Procrustes & CKA & PWCCA & $\bar \rho_{\text{CCA}}$ & $R^2_{\text{CCA}}$ \\
		\midrule
%			1 & 0.718 (1.7E-31) & 0.689 (3.3E-29) & 0.719 (1.4E-31) & 0.724 (1.0E+00) & 0.724 (1.0E+00) \\
%2 & 0.564 (3.7E-20) & 0.535 (2.3E-18) & 0.565 (3.0E-20) & 0.568 (1.0E+00) & 0.568 (1.0E+00) \\
%3 & 0.647 (5.5E-26) & 0.638 (2.6E-25) & 0.592 (5.1E-22) & 0.612 (1.0E+00) & 0.613 (1.0E+00) \\
%4 & 0.489 (1.3E-15) & 0.478 (5.2E-15) & 0.452 (1.4E-13) & 0.409 (1.0E+00) & 0.399 (1.0E+00) \\
%5 & -0.010 (5.6E-01) & 0.036 (2.8E-01) & -0.072 (8.8E-01) & -0.102 (4.9E-02) & -0.116 (3.0E-02) \\
%6 & 0.255 (1.8E-05) & 0.359 (3.1E-09) & 0.189 (1.1E-03) & 0.153 (9.9E-01) & 0.140 (9.9E-01) \\
%7 & 0.184 (1.4E-03) & 0.316 (1.6E-07) & 0.084 (8.7E-02) & 0.094 (9.4E-01) & 0.071 (8.7E-01) \\
%8 & 0.348 (8.8E-09) & 0.420 (5.1E-12) & 0.306 (3.6E-07) & 0.352 (1.0E+00) & 0.316 (1.0E+00) \\
%9 & 0.545 (5.7E-19) & 0.605 (6.4E-23) & 0.525 (9.9E-18) & 0.537 (1.0E+00) & 0.523 (1.0E+00) \\
%10 & 0.593 (4.0E-22) & 0.645 (7.5E-26) & 0.489 (1.2E-15) & 0.563 (1.0E+00) & 0.558 (1.0E+00) \\
%11 & 0.651 (3.1E-26) & 0.687 (4.7E-29) & 0.561 (5.7E-20) & 0.610 (1.0E+00) & 0.608 (1.0E+00) \\
12 & 0.707 (1.2E-30) & 0.732 (1.0E-32) & 0.637 (3.1E-25) & 0.662 (1.0E+00) & 0.658 (1.0E+00) \\
		\bottomrule
	\end{tabular}
\end{table}
 
\clearpage

\subsection{Perturbation: pretraining seed and principal component deletion}
We find that for these experiments, results are consistent across the layers we analyze (the last 6 layers of BERT base). Tables \ref{table:pca-sst2-rho} and \ref{table:pca-sst2-tau} show results for Spearman $\rho$ and Kendall's $\tau$, respectively.

% SST-2 rho PCA deletion
\begin{table}[!h]
\centering
	\caption{Layer-wise Spearman $\rho$ results for perturbing pretraining seed and principal component deletion, and assessing functionality through the SST-2 probe}
	\label{table:pca-sst2-rho}
	\begin{tabular}{cccccc}
		\toprule
		Layer & Procrustes & CKA & PWCCA & $\bar \rho_{\text{CCA}}$ & $R^2_{\text{CCA}}$ \\
		\midrule
			8 & 0.764 (2.4E-36) & 0.668 (3.2E-25) & 0.776 (3.4E-38) & 0.700 (1.9E-28) & 0.700 (1.8E-28) \\
9 & 0.813 (2.1E-44) & 0.706 (4.0E-29) & 0.825 (9.2E-47) & 0.728 (1.3E-31) & 0.728 (1.2E-31) \\
10 & 0.873 (2.1E-58) & 0.818 (2.7E-45) & 0.874 (1.1E-58) & 0.748 (3.2E-34) & 0.749 (2.7E-34) \\
11 & 0.918 (1.2E-74) & 0.797 (1.4E-41) & 0.922 (1.7E-76) & 0.781 (6.6E-39) & 0.781 (7.0E-39) \\
12 & 0.932 (1.1E-81) & 0.766 (1.1E-36) & 0.955 (4.2E-97) & 0.810 (6.1E-44) & 0.810 (6.1E-44) \\
		\bottomrule
	\end{tabular}
\end{table}

% SST-2 tau PCA deletion
\begin{table}[h]
\centering
	\caption{Layer-wise Kendall's $\tau$ results for perturbing pretraining seed and principal component deletion, and assessing functionality through the SST-2 probe}
	\label{table:pca-sst2-tau}
	\begin{tabular}{cccccc}
		\toprule
		Layer & Procrustes & CKA & PWCCA & $\bar \rho_{\text{CCA}}$ & $R^2_{\text{CCA}}$ \\
		\midrule
			8 & 0.560 (1.8E-29) & 0.479 (4.4E-22) & 0.573 (1.1E-30) & 0.512 (6.8E-25) & 0.512 (6.6E-25) \\
9 & 0.602 (1.2E-33) & 0.509 (1.2E-24) & 0.618 (2.5E-35) & 0.542 (1.1E-27) & 0.543 (9.7E-28) \\
10 & 0.684 (5.6E-43) & 0.627 (2.1E-36) & 0.685 (5.3E-43) & 0.588 (2.9E-32) & 0.589 (2.5E-32) \\
11 & 0.751 (2.8E-51) & 0.616 (3.3E-35) & 0.756 (6.4E-52) & 0.648 (9.2E-39) & 0.648 (9.2E-39) \\
12 & 0.787 (3.4E-56) & 0.588 (2.9E-32) & 0.819 (1.2E-60) & 0.701 (4.7E-45) & 0.701 (4.9E-45) \\
		\bottomrule
	\end{tabular}
\end{table}

\clearpage
\subsection{Perturbation: fine-tuning seed, Functionality: HANS} \label{appendix:feathertable}
Results for this experiment are similar across layers for Procrustes and all three CCA-based measures, with middle layers of BERT base having a slightly higher rank correlation score in general. For CKA, this effect is even more pronounced. Tables \ref{table:feather-rho} and \ref{table:feather-tau} show the results for Spearman $\rho$ and Kendall's $\tau$, respectively.

% feather rho 
\begin{table}[h]
\centering
	\caption{Layer-wise Spearman $\rho$ results for perturbing finetuning seed, and assessing functionality through the HANS: Lexical (non-entailment) OOD dataset}
	\label{table:feather-rho}
	\begin{tabular}{cccccc}
		\toprule
		Layer & Procrustes ($p$) & CKA ($p$) & PWCCA ($p$) & $\bar \rho_{\text{CCA}}$ ($p$) & $R^2_{\text{CCA}}$ ($p$)\\
		\midrule
			1 & 0.425 (5.1E-06) & 0.361 (1.1E-04) & 0.405 (1.4E-05) & 0.388 (3.4E-05) & 0.389 (3.2E-05) \\
			2 & 0.510 (3.1E-08) & 0.410 (1.2E-05) & 0.486 (1.5E-07) & 0.488 (1.3E-07) & 0.483 (1.8E-07) \\
			3 & 0.531 (6.6E-09) & 0.427 (4.6E-06) & 0.538 (3.8E-09) & 0.533 (5.6E-09) & 0.532 (6.2E-09) \\
			4 & 0.543 (2.6E-09) & 0.506 (3.9E-08) & 0.552 (1.4E-09) & 0.555 (1.0E-09) & 0.550 (1.5E-09) \\
			5 & 0.563 (5.3E-10) & 0.512 (2.6E-08) & 0.570 (2.9E-10) & 0.582 (1.1E-10) & 0.580 (1.3E-10) \\
			6 & 0.629 (1.2E-12) & 0.641 (3.6E-13) & 0.621 (2.8E-12) & 0.621 (2.7E-12) & 0.622 (2.5E-12) \\
			7 & 0.647 (1.7E-13) & 0.658 (5.0E-14) & 0.647 (1.7E-13) & 0.653 (9.0E-14) & 0.650 (1.2E-13) \\
			8 & 0.643 (2.7E-13) & 0.552 (1.3E-09) & 0.653 (9.5E-14) & 0.651 (1.1E-13) & 0.651 (1.2E-13) \\
			9 & 0.589 (5.9E-11) & 0.419 (7.1E-06) & 0.641 (3.5E-13) & 0.662 (3.3E-14) & 0.660 (4.2E-14) \\
			10 & 0.536 (4.6E-09) & 0.437 (2.7E-06) & 0.559 (7.3E-10) & 0.612 (6.6E-12) & 0.614 (5.4E-12) \\
			11 & 0.532 (6.2E-09) & 0.426 (4.9E-06) & 0.565 (4.7E-10) & 0.619 (3.4E-12) & 0.614 (5.5E-12) \\
			12 & 0.465 (5.3E-07) & 0.192 (2.8E-02) & 0.574 (2.1E-10) & 0.609 (9.2E-12) & 0.610 (7.9E-12) \\
		\bottomrule
	\end{tabular}
\end{table}

% feather tau
\begin{table}[h]
\centering
	\caption{Layer-wise Kendall's $\tau$ results for perturbing finetuning seed, and assessing functionality through the HANS: Lexical (non-entailment) OOD dataset}
	\label{table:feather-tau}
	\begin{tabular}{cccccc}
		\toprule
		Layer & Procrustes ($p$) & CKA ($p$) & PWCCA ($p$) & $\bar \rho_{\text{CCA}}$ ($p$) & $R^2_{\text{CCA}}$ ($p$)\\
		\midrule
			1 & 0.295 (6.7E-06) & 0.269 (3.6E-05) & 0.277 (2.2E-05) & 0.265 (4.7E-05) & 0.268 (4.0E-05) \\
			2 & 0.363 (4.6E-08) & 0.288 (1.1E-05) & 0.343 (2.1E-07) & 0.342 (2.3E-07) & 0.342 (2.4E-07) \\
			3 & 0.372 (2.1E-08) & 0.290 (9.5E-06) & 0.378 (1.3E-08) & 0.375 (1.6E-08) & 0.375 (1.6E-08) \\
			4 & 0.393 (3.4E-09) & 0.358 (6.6E-08) & 0.401 (1.7E-09) & 0.405 (1.2E-09) & 0.403 (1.4E-09) \\
			5 & 0.410 (7.7E-10) & 0.367 (3.3E-08) & 0.417 (4.1E-10) & 0.428 (1.4E-10) & 0.424 (2.0E-10) \\
			6 & 0.464 (4.2E-12) & 0.474 (1.5E-12) & 0.460 (6.3E-12) & 0.460 (5.8E-12) & 0.461 (5.6E-12) \\
			7 & 0.483 (5.5E-13) & 0.488 (3.3E-13) & 0.481 (7.1E-13) & 0.486 (3.9E-13) & 0.483 (5.5E-13) \\
			8 & 0.478 (9.2E-13) & 0.392 (3.7E-09) & 0.483 (5.7E-13) & 0.481 (6.5E-13) & 0.480 (7.7E-13) \\
			9 & 0.432 (1.0E-10) & 0.293 (7.7E-06) & 0.475 (1.2E-12) & 0.496 (1.3E-13) & 0.494 (1.6E-13) \\
			10 & 0.380 (1.0E-08) & 0.306 (3.4E-06) & 0.401 (1.7E-09) & 0.447 (2.3E-11) & 0.448 (2.1E-11) \\
			11 & 0.376 (1.5E-08) & 0.292 (8.3E-06) & 0.411 (6.9E-10) & 0.448 (2.1E-11) & 0.445 (2.7E-11) \\
			12 & 0.330 (5.7E-07) & 0.127 (3.1E-02) & 0.416 (4.4E-10) & 0.446 (2.5E-11) & 0.447 (2.2E-11) \\
		\bottomrule
	\end{tabular}
\end{table}

\subsection{Perturbation: pretraining seeds and finetuning seeds of BERT medium}

Rank correlation scores are low across the board for this task, suggesting that it is difficult for all existing dissimilarity measures, regardless of the layer within a network. Results on the Antonymy stress test for Spearman $\rho$ and Kendall's $\tau$ are in Tables \ref{table:medium-antonymy-rho} and \ref{table:medium-antonymy-tau}, respectively. Results on the Numerical stress test for Spearman $\rho$ and Kendall's $\tau$ are in Tables \ref{table:medium-numerical-rho} and \ref{table:medium-numerical-tau}, respectively. 

% Antonymy rho
\begin{table}
\centering
	\caption{Layer-wise Spearman $\rho$ results for perturbing pretraining seed and finetuning seed, and assessing functionality through the Antonymy stress test}
	\label{table:medium-antonymy-rho}
	\begin{tabular}{cccccc}
		\toprule
		Layer & Procrustes & CKA & PWCCA & $\bar \rho_{\text{CCA}}$ & $R^2_{\text{CCA}}$ \\
		\midrule
			1 & 0.252 (5.7E-03) & 0.241 (7.8E-03) & 0.168 (4.7E-02) & 0.305 (1.0E+00) & 0.327 (1.0E+00) \\
2 & 0.213 (1.7E-02) & 0.145 (7.5E-02) & 0.131 (9.7E-02) & 0.047 (6.8E-01) & 0.031 (6.2E-01) \\
3 & 0.260 (4.5E-03) & 0.262 (4.2E-03) & 0.208 (1.9E-02) & 0.137 (9.1E-01) & 0.111 (8.6E-01) \\
4 & 0.260 (4.5E-03) & 0.265 (3.8E-03) & 0.265 (3.8E-03) & 0.276 (1.0E+00) & 0.254 (9.9E-01) \\
5 & 0.273 (3.0E-03) & 0.302 (1.1E-03) & 0.278 (2.5E-03) & 0.339 (1.0E+00) & 0.310 (1.0E+00) \\
6 & 0.330 (3.9E-04) & 0.280 (2.4E-03) & 0.346 (2.1E-04) & 0.313 (1.0E+00) & 0.304 (1.0E+00) \\
7 & 0.271 (3.2E-03) & 0.315 (7.1E-04) & 0.111 (1.4E-01) & 0.091 (8.2E-01) & 0.090 (8.1E-01) \\
8 & 0.084 (2.0E-01) & 0.004 (4.8E-01) & 0.123 (1.1E-01) & 0.204 (9.8E-01) & 0.198 (9.8E-01) \\
		\bottomrule
	\end{tabular}
\end{table}

% Antonymy tau
\begin{table}
\centering
	\caption{Layer-wise Kendall's $\tau$ results for perturbing pretraining seed and finetuning seed, and assessing functionality through the Antonymy stress test}
	\label{table:medium-antonymy-tau}
	\begin{tabular}{cccccc}
		\toprule
		Layer & Procrustes & CKA & PWCCA & $\bar \rho_{\text{CCA}}$ & $R^2_{\text{CCA}}$ \\
		\midrule
			1 & 0.199 (1.7E-03) & 0.171 (5.9E-03) & 0.126 (3.3E-02) & 0.244 (1.0E+00) & 0.243 (1.0E+00) \\
2 & 0.179 (4.3E-03) & 0.123 (3.5E-02) & 0.118 (4.2E-02) & 0.061 (8.1E-01) & 0.042 (7.3E-01) \\
3 & 0.185 (3.3E-03) & 0.186 (3.2E-03) & 0.139 (2.0E-02) & 0.110 (9.5E-01) & 0.096 (9.2E-01) \\
4 & 0.187 (3.0E-03) & 0.191 (2.6E-03) & 0.188 (2.9E-03) & 0.206 (1.0E+00) & 0.193 (1.0E+00) \\
5 & 0.192 (2.4E-03) & 0.194 (2.2E-03) & 0.202 (1.5E-03) & 0.267 (1.0E+00) & 0.242 (1.0E+00) \\
6 & 0.236 (2.7E-04) & 0.197 (1.9E-03) & 0.252 (1.1E-04) & 0.229 (1.0E+00) & 0.221 (1.0E+00) \\
7 & 0.189 (2.8E-03) & 0.217 (7.3E-04) & 0.091 (9.1E-02) & 0.081 (8.8E-01) & 0.082 (8.9E-01) \\
8 & 0.061 (1.9E-01) & -0.000 (5.0E-01) & 0.101 (6.9E-02) & 0.155 (9.9E-01) & 0.150 (9.9E-01) \\
		\bottomrule
	\end{tabular}
\end{table}

% Numerical rho
\begin{table}
\centering
	\caption{Layer-wise Spearman $\rho$ results for perturbing pretraining seed and finetuning seed, and assessing functionality through the Numerical stress test}
	\label{table:medium-numerical-rho}
	\begin{tabular}{cccccc}
		\toprule
		Layer & Procrustes & CKA & PWCCA & $\bar \rho_{\text{CCA}}$ & $R^2_{\text{CCA}}$ \\
		\midrule
			1 & 0.137 (8.7E-02) & 0.108 (1.4E-01) & 0.107 (1.4E-01) & 0.072 (7.6E-01) & 0.072 (7.6E-01) \\
2 & -0.012 (5.5E-01) & 0.060 (2.8E-01) & 0.062 (2.7E-01) & 0.004 (5.1E-01) & 0.001 (5.0E-01) \\
3 & -0.059 (7.2E-01) & 0.011 (4.6E-01) & -0.031 (6.2E-01) & -0.060 (2.8E-01) & -0.056 (2.9E-01) \\
4 & 0.041 (3.4E-01) & 0.052 (3.0E-01) & -0.026 (6.0E-01) & -0.101 (1.6E-01) & -0.084 (2.0E-01) \\
5 & 0.003 (4.9E-01) & 0.131 (9.7E-02) & -0.047 (6.8E-01) & -0.061 (2.7E-01) & -0.061 (2.7E-01) \\
6 & 0.092 (1.8E-01) & 0.260 (4.5E-03) & -0.029 (6.1E-01) & -0.064 (2.6E-01) & -0.056 (2.9E-01) \\
7 & 0.164 (5.2E-02) & 0.250 (6.1E-03) & 0.037 (3.6E-01) & 0.040 (6.5E-01) & 0.040 (6.5E-01) \\
8 & 0.202 (2.2E-02) & 0.105 (1.5E-01) & 0.175 (4.1E-02) & 0.134 (9.1E-01) & 0.143 (9.2E-01) \\
		\bottomrule
	\end{tabular}
\end{table}

% Numerical tau
\begin{table}
\centering
	\caption{Layer-wise Kendall's $\tau$ results for perturbing pretraining seed and finetuning seed, and assessing functionality through the Numerical stress test}
	\label{table:medium-numerical-tau}
	\begin{tabular}{cccccc}
		\toprule
		Layer & Procrustes & CKA & PWCCA & $\bar \rho_{\text{CCA}}$ & $R^2_{\text{CCA}}$ \\
		\midrule
			1 & 0.103 (6.5E-02) & 0.083 (1.1E-01) & 0.074 (1.4E-01) & 0.050 (7.7E-01) & 0.048 (7.6E-01) \\
2 & -0.010 (5.6E-01) & 0.046 (2.5E-01) & 0.046 (2.5E-01) & 0.006 (5.3E-01) & 0.001 (5.0E-01) \\
3 & -0.041 (7.3E-01) & 0.014 (4.2E-01) & -0.018 (6.0E-01) & -0.047 (2.5E-01) & -0.047 (2.4E-01) \\
4 & 0.031 (3.2E-01) & 0.038 (2.9E-01) & -0.020 (6.2E-01) & -0.076 (1.3E-01) & -0.065 (1.7E-01) \\
5 & 0.005 (4.7E-01) & 0.086 (1.0E-01) & -0.031 (6.8E-01) & -0.042 (2.7E-01) & -0.042 (2.7E-01) \\
6 & 0.060 (1.9E-01) & 0.175 (5.1E-03) & -0.020 (6.2E-01) & -0.050 (2.3E-01) & -0.046 (2.5E-01) \\
7 & 0.112 (4.9E-02) & 0.168 (6.8E-03) & 0.030 (3.3E-01) & 0.019 (6.1E-01) & 0.024 (6.4E-01) \\
8 & 0.131 (2.7E-02) & 0.063 (1.8E-01) & 0.125 (3.3E-02) & 0.099 (9.3E-01) & 0.103 (9.4E-01) \\
		\bottomrule
	\end{tabular}
\end{table}

\clearpage

%% file: cifar10c_all_table.tex
\begin{table}[h]
\caption{Results for perturbing training seed and assessing functionality through CIFAR-10C}
\begin{minipage}{.5\linewidth}
\centering
\caption{Spearman $\rho$ results}
	\label{table:cifar10c-rho}
	\begin{tabular}{lrr}
		\toprule
		Corruption & Procrustes & CKA \\
		\midrule
gaussian\_noise	&	0.083	&	0.076	\\
shot\_noise	&	0.171	&	0.161	\\
impulse\_noise	&	0.104	&	0.083	\\
defocus\_blur	&	-0.025	&	0.021	\\
glass\_blur	&	0.082	&	0.073	\\
motion\_blur	&	0.033	&	0.035	\\
zoom\_blur	&	-0.023	&	0.020\\
snow	&	0.087	&	0.060\\
frost	&	-0.062	&	-0.081	\\
fog	&	-0.029	&	-0.039	\\
brightness	&	0.122	&	0.110\\
contrast	&	-0.225	&	-0.145	\\
elastic\_transform	&	0.137	&	0.122	\\
pixelate	&	0.118	&	0.098	\\
jpeg\_compression	&	0.149	&	0.102	\\
speckle\_noise	&	0.028	&	0.033	\\
gaussian\_blur	&	0.149	&	0.141	\\
spatter	&	0.089	&	0.079	\\
saturate	&	0.143	&	0.135	\\
\midrule
Average	&	0.060	&	0.057	\\
		\bottomrule
	\end{tabular}
\end{minipage}%
    \begin{minipage}{.5\linewidth}	
    \centering
\caption{Kendall $\tau$ results  }
	\label{table:cifar10c-tau}
	\begin{tabular}{lrr}
		\toprule
		Corruption & Procrustes & CKA \\
		\midrule
gaussian\_noise	&	0.057	&	0.050	\\
shot\_noise	&	0.118	&	0.110 \\
impulse\_noise	&	0.070&	0.055	\\
defocus\_blur	&	-0.016	&	0.013	\\
glass\_blur	&	0.057	&	0.047	\\
motion\_blur	&	0.021	&	0.022	\\
zoom\_blur	&	-0.014	&	0.013	\\
snow	&	0.059	&	0.042	\\
frost	&	-0.046	&	-0.059	\\
fog	&	-0.020	&	-0.025	\\
brightness	&	0.084	&	0.077	\\
contrast	&	-0.158	&	-0.102	\\
elastic\_transform	&	0.094	&	0.085	\\
pixelate	&	0.081	&	0.066	\\
jpeg\_compression	&	0.103	&	0.070\\
speckle\_noise	&	0.019	&	0.022	\\
gaussian\_blur	&	0.102	&	0.095	\\
spatter	&	0.059	&	0.053	\\
saturate	&	0.100	&	0.096	\\
\midrule
Average	&	0.041	&	0.038	\\
		\bottomrule
	\end{tabular}
	    \end{minipage}

\end{table}